\documentclass{article}

\usepackage[final]{neurips_2025}

\usepackage[T1]{fontenc}
\usepackage[utf8]{inputenc}       % 用 pdflatex 的话保留
\usepackage{hyperref}       % hyperlinks
\usepackage{url}            % simple URL typesetting
\usepackage{booktabs}       % professional-quality tables
\usepackage{amsfonts}       % blackboard math symbols
\usepackage{nicefrac}       % compact symbols for 1/2, etc.
\usepackage{microtype}      % microtypography
\usepackage{xcolor}         % colors
\usepackage{microtype}
\usepackage{graphicx}
\usepackage{tabularx}   % for the X column type
\usepackage{amsmath}
\usepackage{amssymb}
\usepackage{mathtools}
\usepackage{amsthm}
\usepackage[capitalize,noabbrev]{cleveref}
\usepackage{CJKutf8} % 引入CJKutf8宏包
\usepackage{subcaption}
\usepackage{graphicx}
\usepackage{multirow}
\usepackage{array}
\DeclareGraphicsExtensions{.pdf,.png,.jpg}
\newtheorem{defi}{Definition}

\title{Hierarchical Frequency Tagging Probe (HFTP): A Unified Approach to Investigate Syntactic Structure Representations in Large Language Models and the Human Brain}

\author{%
  \parbox{\textwidth}{\centering\bfseries
    Jingmin An$^{1}$, Yilong Song$^{1}$, Ruolin Yang$^{1}$, Nai Ding$^{2}$, Lingxi Lu$^{3}$\\
    Yuxuan Wang$^{1}$, Wei Wang$^{4}$, Chu Zhuang$^{1}$, Qian Wang$^{1,\dagger}$, Fang Fang$^{1,\dagger}$%
  }\\[6pt]
  \parbox{\textwidth}{\centering
    $^{1}$Peking University, $^{2}$Zhejiang University, $^{3}$Beijing Language and Culture University,\\
    $^{4}$Beijing Institute for General Artificial Intelligence
  }\\
  \texttt{anjm@stu.pku.edu.cn, \{wangqianpsy, ffang\}@pku.edu.cn}%
}

\begin{document}

\maketitle

\begin{abstract}
\label{abstract}
  Large Language Models (LLMs) demonstrate human-level or even superior language abilities, effectively modeling syntactic structures, yet the specific computational modules responsible remain unclear. A key question is whether LLM behavioral capabilities stem from mechanisms akin to those in the human brain. To address these questions, we introduce the Hierarchical Frequency Tagging Probe (HFTP), a tool that utilizes frequency-domain analysis to identify neuron-wise components of LLMs (e.g., individual Multilayer Perceptron (MLP) neurons) and cortical regions (via intracranial recordings) encoding syntactic structures. Our results show that models such as GPT-2, Gemma, Gemma 2, Llama 2, Llama 3.1, and GLM-4 process syntax in analogous layers, while the human brain relies on distinct cortical regions for different syntactic levels. Representational similarity analysis reveals a stronger alignment between LLM representations and the left hemisphere of the brain (dominant in language processing). Notably, upgraded models exhibit divergent trends: Gemma 2 shows greater brain similarity than Gemma, while Llama 3.1 shows less alignment with the brain compared to Llama 2. These findings offer new insights into the interpretability of LLM behavioral improvements, raising questions about whether these advancements are driven by human-like or non-human-like mechanisms, and establish HFTP as a valuable tool bridging computational linguistics and cognitive neuroscience. This project is available at \url{https://github.com/LilTiger/HFTP}.
\end{abstract}

\section{Introduction}
\label{introduction}
Language is fundamental to human communication, thought, and cultural transmission. According to the framework proposed by Noam Chomsky, language is divided into three key components: semantics (meaning), phonology (sound), and syntax (hierarchical sentence structure) \citep{chomsky1965}. Syntax is particularly crucial as it governs how words combine into meaningful expressions, underpinning the recursive and generative capacity unique to human language. The theory of Universal Grammar proposes that all human languages share innate structural principles \citep{chomsky1980rules}. Building on this foundation, cognitive neuroscience has shown that syntactic processing recruits mechanisms distinct from other linguistic functions, particularly within the left inferior frontal and posterior temporal regions \citep{matchin2020syntax,friederici2011brain}.  Moreover, increases in syntactic complexity yield graded activation in left inferior frontal and posterior temporal cortices, consistent with the computation and maintenance of hierarchical dependencies rather than simple lexical associations \citep{pallier2011cortical,nelson2017neurophysiological}. As our understanding of human syntactic computation deepens, artificial intelligence models have increasingly sought to emulate this ability to capture and represent structured language.

In recent years, large language models (LLMs) have evolved rapidly, achieving human-level or better performance on a range of linguistic benchmarks and professional exams \citep{achiam2023gpt}. Their success in understanding, translation, and summarization has led to claims of human-like fluency, particularly in generating text that conforms to surface syntactic regularities \citep{mahowald2024dissociating,he2024exploring,van2024adapted}. Yet it remains unclear whether such models truly represent the hierarchical sentence structures that characterize human syntax. Some findings indicate that LLMs can implicitly capture and manipulate structural relations \citep{manning2020emergent}, while others suggest their success can rely on shallow statistical heuristics rather than genuine structural understanding \citep{linzen2016assessing,mccoy2019right}. This ongoing debate underscores the need for a unified analytic framework capable of directly comparing syntactic representations in human and model systems, which is essential for probing the depth and nature of syntactic alignment between the human brain and artificial models.

Ding et al. \citep{dingCorticalTrackingHierarchical2016} introduced the hierarchical frequency tagging (HFT) technique to uncover how the human brain processes hierarchical linguistic structures during natural speech comprehension. In this paradigm, monosyllabic words are presented at a rate of \(4\;\mathrm{Hz}\) to form phrases at \(2\;\mathrm{Hz}\), which combine into sentences at \(1\;\mathrm{Hz}\). Using frequency-domain analysis of electrophysiological signals, Ding et al. deconstruct the processing of linguistic structures such as phrases and sentences. Subsequent work has extended the HFT framework along complementary axes. Attention is required to group lower-level inputs into higher-order linguistic units, and diverting attention attenuates word- and phrase-rate tracking \citep{ding2018attention}. MEG source analyses dissociate cortical signatures for word- versus phrase-level rhythms and link phrasal tracking to comprehension \citep{keitel2018perceptually}. Computational modelling shows that oscillatory architectures can implement hierarchical parsing and reproduce HFT-like spectra \citep{martin2017mechanism}. Natural-speech EEG reveals endogenous word-rate tracking that interacts with exogenous rhythmic cues \citep{luo2020cortical}. Naturalistic experiments further indicate that phrase-rate tracking indexes internally generated structure rather than compositional meaning per se, while remaining sensitive to lexical–syntactic information that enables structure building \citep{coopmans2022effects}. These studies demonstrate the effectiveness of HFT for isolating neural markers of hierarchical language processing. Accordingly, HFTP leverages frequency-domain tagging to separate sentence- and phrase-level structure from lexical and prosodic regularities, and to yield robust spectral markers that can be aligned quantitatively with language-model representations across timescales, thereby providing a principled bridge between biological and artificial systems.

Building on the HFT paradigm \cite{dingCorticalTrackingHierarchical2016}, here we developed the Hierarchical Frequency Tagging Probe (HFTP) to investigate whether specific computational modules within LLMs process hierarchical sentence structures. HFTP offers a unified approach to explore internal similarities and systematically examine the alignment of syntactic structure representations between LLMs and the human brain. The key contributions of this paper are: \textbf{(i)}  We innovatively employed frequency-domain analysis using HFTP to characterize the syntactic structure representations of every computational module in each layer of LLMs; \textbf{(ii)} HFTP provides a simple, universally applicable approach for detecting and aligning syntactic structure representations in LLMs (via neuron-wise probing) and the human brain (via population-level analyses), and extends seamlessly to naturalistic text. \textbf{ (iii) }Using syntactic templates derived from HFTP, we identified brain regions highly correlated with LLMs, predominantly located in key language-processing areas of the left hemisphere; \textbf{(iv)} By comparing six LLMs, we observed divergent trends in upgraded versions, with some showing increased similarity to brain representations while others exhibited reduced alignment. In sum, HFTP effectively detects syntactic structure representations in both LLMs and the human brain, providing a novel framework for alignment study.

\section{Related work}

\textbf{Syntactic processing in the human brain}
In humans, syntactic processing recruits a left-dominant fronto-temporal network that supports hierarchical combination from finite elements. Classic lesion and neuroimaging work documents a left-hemisphere advantage \citep{friederici2009complexity,Hagoort2013,blank2016syntactic}, with hemispheric temporal sensitivities aiding speech segmentation \citep{albouy2020distinct}. Converging evidence shows that syntactic operations are distributed across frontal and temporal cortex with substantial overlap with semantic integration \citep{blank2016syntactic,fedorenko2020lack}. Artificial-grammar fMRI further indicates that hierarchically structured strings reliably engage left inferior frontal gyrus and posterior superior temporal regions \citep{chen2021hierarchical}. Overall, these findings demonstrate that syntactic processing is not confined to isolated regions but is part of a broad, interconnected network.

\textbf{Syntactic processing in language models}
Even before the development of LLMs, researchers found that simple LSTM language models could capture syntax-sensitive dependencies, such as subject-verb agreement \citep{linzen2016assessing, kuncoro2018lstms}. Using a technique called structural probing, Manning and colleagues discovered that transformer-based models like BERT can encode hierarchical syntactic trees, enabling such models to implicitly represent complex syntax without direct training \citep{hewittStructuralProbeFinding2019}. These transformer-based models excel at tracking both local and long-range dependencies through specialized attention mechanisms, distributing syntactic knowledge across layers \citep{clark2019does, tenney2019bert, manning2020emergent}.However, the methods employed in these studies of language models make it challenging to apply findings to the exploration of human brain activity.

\textbf{Alignment between LLMs and the human brain}
A growing body of work shows that sentence-level \emph{contextual} embeddings from predictive LMs strongly predict cortical responses during comprehension \citep{sun2020neural,Schrimpf2021}. Disentanglement analyses that factorize activations into lexical, compositional, syntactic, and semantic components indicate distributed contributions—often with compositional/lexical signals explaining much of the alignment, rather than syntax alone \citep{caucheteux2021disentangling,caucheteux2022brains}. Neuroimaging manipulations that dissociate semantics from syntax reveal distinct patterns (including frontal engagement) without establishing syntactic dominance \citep{wang2020probing}. Intervention studies that selectively remove linguistic properties from model representations yield reliable drops in brain alignment, with syntactic properties (e.g., tree depth, top constituents) exerting large cross-layer effects \citep{oota2023joint}. These findings are consistent with convergence on shared representational axes across brains and models \citep{hosseini2024universality} and with demonstrations that model-derived stimuli can causally drive or suppress the human language network \citep{tuckute2024driving}, while alignment varies systematically across layers as contextual information accrues \citep{goldstein2022shared}. 
However, despite these advances, methodological inconsistencies still prevent systematic comparison of syntactic structure encoding across models and neural populations, underscoring the need for a unified analytical framework.

\begin{figure*}[ht]
\centering
\includegraphics[width=\linewidth]{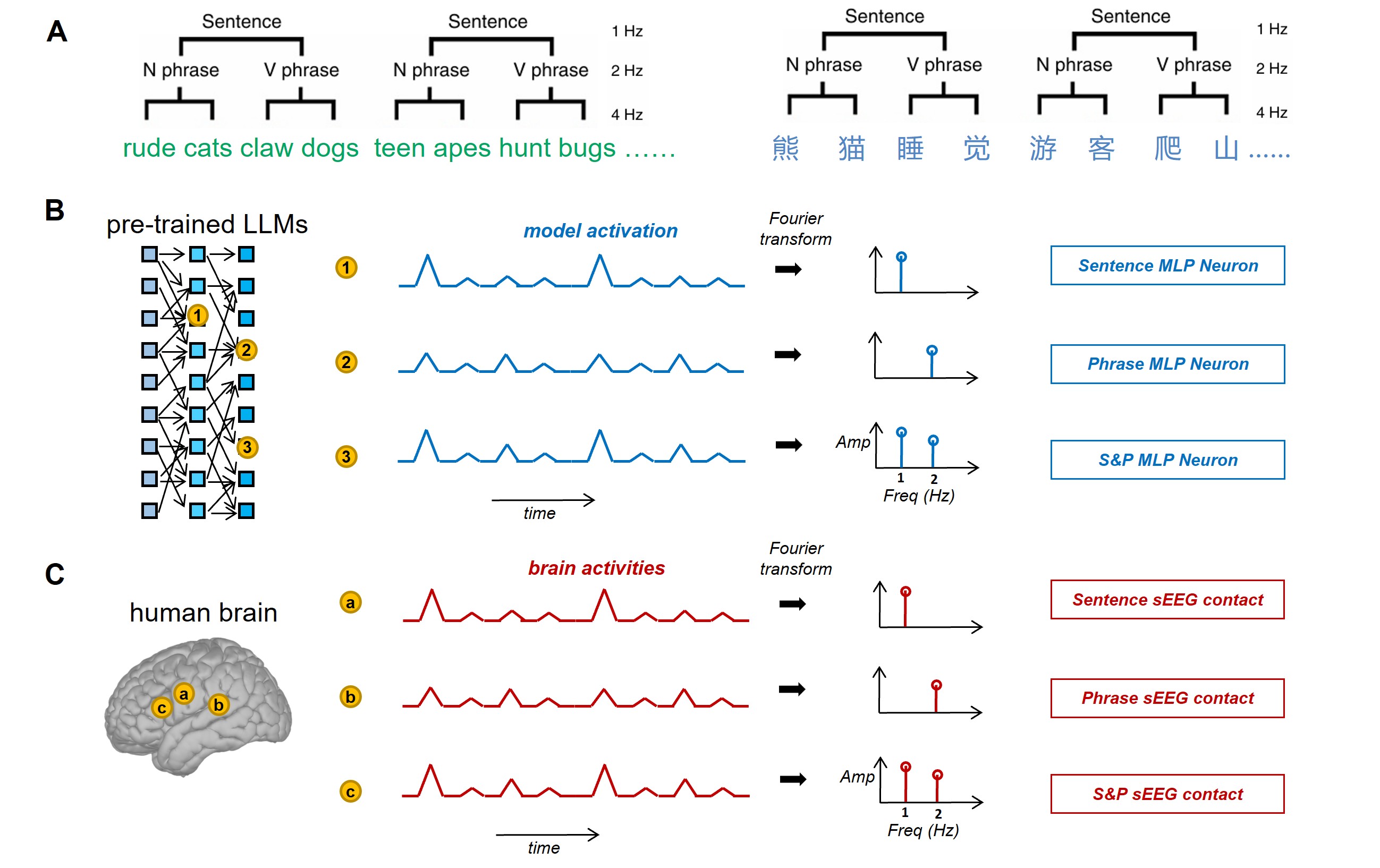}
\captionsetup{aboveskip=-2pt} % 缩小caption和图片的距离
\caption{A framework for Hierarchical Frequency Tagging Probes (HFTP) and an illustration of neurons involved in different levels of hierarchical linguistic processing in both LLMs and the human brain. \textbf{A}, hierarchical linguistic structure in English and Chinese including syllable, phrase, and sentence. \textbf{B}, hierarchical linguistic pattern (\(1\;\mathrm{Hz}\): sentence feature, \(2\;\mathrm{Hz}\): phrase feature) observed both in LLMs and \textbf{C},  human brain.}
    \label{framework}
\vskip -0.2in
\end{figure*}

\section{Methods}

\begin{figure*}
\centering
\includegraphics[width=\linewidth]{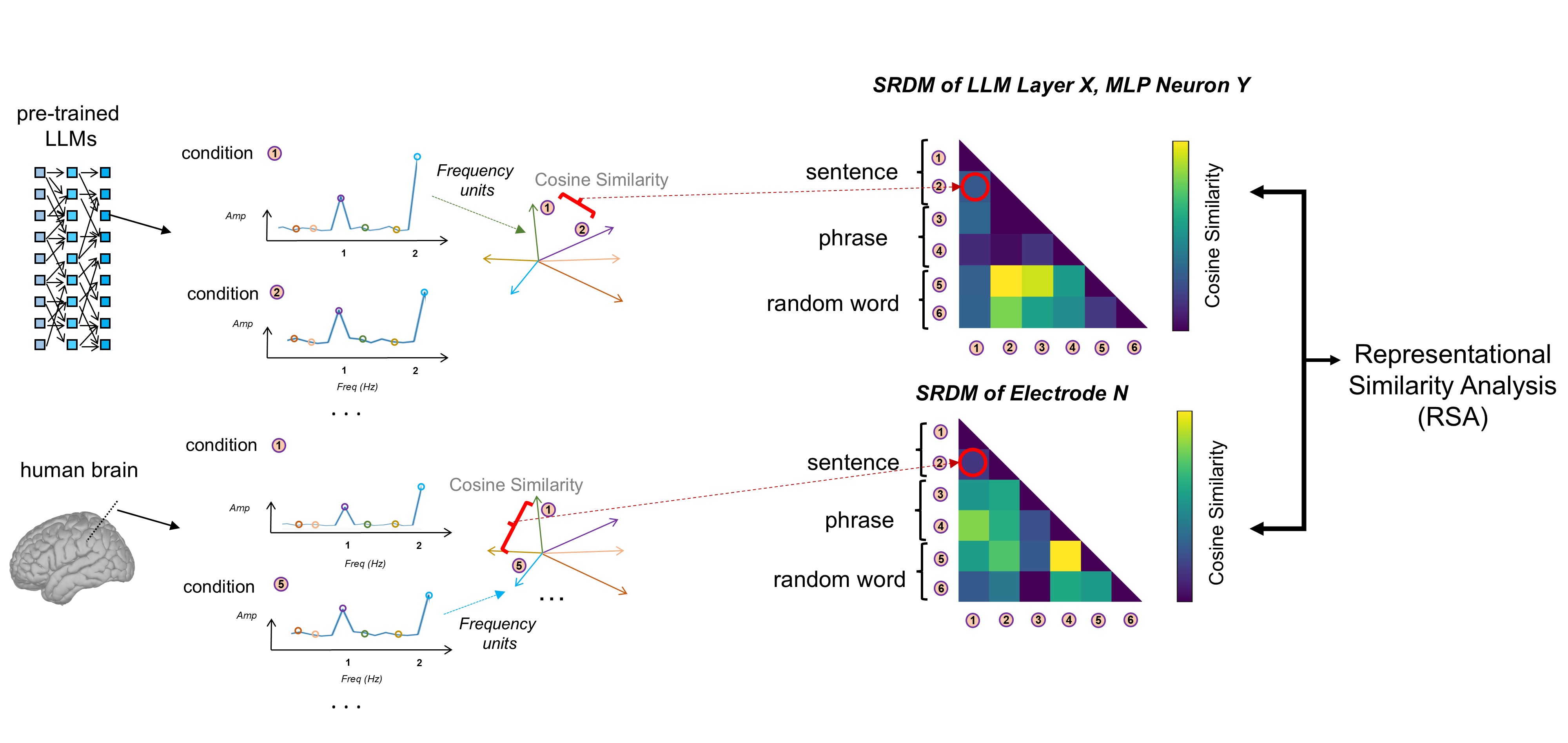}
\captionsetup{aboveskip=-2pt}
\caption{Alignment pipeline between LLMs and the human brain. SRDMs are computed for exclusive sentence/phrase and sentence\&phrase MLP neurons and brain channels by comparing cosine similarities across different conditions. Subsequently, RSA (using Spearman correlation) is applied to quantify the similarity between the two SRDMs, thereby assessing the correspondence between model and brain representations.}
\label{searchlight}
\vskip -0.15in
\end{figure*}

We present the framework of the proposed HFTP methodology (see Figure \ref{framework}). This framework is organized into four parts: Section \ref{data_llm} describes the syntactic corpora used and the LLM architectures; Section \ref{process_llm} details the application of HFTP to detect significant sentence and phrase neurons; Section \ref{process_human} explains how the HFTP approach is applied to human intracranial stereo-electroencephalography (sEEG) data; and Section \ref{alignment_} correlates syntactic structure representations in LLMs and the human brain by comparing frequency-domain representations and detecting similarities in how syntactic structures are encoded across both systems.

\subsection{Data and LLMs}
\label{data_llm}
We mainly utilized Chinese and English corpora adapted from \citep{dingCorticalTrackingHierarchical2016}, consisting of four-syllable sequences in Chinese or four-word sequences in English, where the first two and last two units form phrases (see Figure \ref{framework}). Further details regarding the corpus can be found in Appendix \ref{corpus}. We also adopt naturalistic text to test the generalizability of our HFTP method (see \ref{natural}). For both the sEEG and model–brain alignment experiments, we used the same two Chinese corpora—the \textbf{sentence} and \textbf{phrase} corpora—from \citep{sheng2019cortical}. While these corpora share a similar structure to the Chinese syntactic corpus used in the LLM experiments, they differ in content. Guided by evidence that periodic lexical regularities alone can produce peaks at word-, phrase-, and sentence-rate frequencies \citep{frank2018lexical}, we added a within-sentence word-order–randomized control in all experiments: items are permuted across positions so that lexical and part-of-speech categories do not recur at fixed positions and any consistent phrase- or sentence-level patterns are prevented, while the lexical set is preserved. This control isolates the syntactic and lexical contribution to \(1\), \(2\), and \(4\;\mathrm{Hz}\) power.

We applied HFTP to six state-of-the-art LLMs: GPT-2, Gemma, Gemma 2, Llama 2, Llama 3.1, and GLM-4, which vary in both architecture and parameter scale (see Table \ref{model_description}). To avoid tokenisation artefacts, we average MLP activations over sub-tokens at the syllable level for Chinese and word level for English before FFT. This ensures that the \(1\;\mathrm{Hz}\) and \(2\;\mathrm{Hz}\) spectral components reflect linguistic boundaries rather than tokenisation boundaries, enabling consistent cross-lingual and cross-model comparisons. Notably, the term “MLP neuron” denotes a computational unit in the intermediate hidden layer of the MLP sub-layer within a Transformer model. This sub-layer consists of two linear transformations separated by a nonlinear activation function. We target MLP neurons because they house localised, interpretable units—conceptualised as “knowledge neurons” \citep{dai2021knowledge} that causally control factual recall and lexical–syntactic concepts \citep{geva2022transformer}. This mechanistic specificity provides the discrete, concept-aware handles required by HFTP to robustly localise syntactic structures.

\subsection{Syntactic structure probe in LLMs}
\label{process_llm}

For each LLM, sequences from both the Chinese and English syntactic corpora were concatenated into a continuous text to capture neural-like activations. During this process, each Chinese syllable (or English word) outputs an activation value, allowing the signal corresponding to every individual linguistic unit to be traced. These time-domain activations were then transformed into frequency-domain information via fast-fourier transform (FFT).  Due to the lack of time-course information inherent to LLM input structures, we artificially  defined a time scale on which the activation values are output at a frequency of \(4\;\mathrm{Hz}\), and we also manually constrained the sampling rate to \(4\;\mathrm{Hz}\), limiting the observable frequencies to the \(0-2\;\mathrm{Hz}\) range. This adjustment ensured that the syntactic rhythms analogous to those observed in human brain data could be captured within the model activations.

LLMs, with their multiple layers and thousands of MLP neurons per layer, require a systematic approach to detect which neurons are responsible for either sentence or phrase processing. HFTP introduces a method to detect significant syntactic processing units, applicable to both LLMs and human brain data. For the LLMs, we conducted a permutation test, randomizing the model activations derived from the structured input corpus 1000 times. The original frequency bins at \(1\;\mathrm{Hz}\) and \(2\;\mathrm{Hz}\), representing sentence and phrase rhythms respectively (their real parts of amplitudes are denoted as $\mathrm{real} [\mathrm{amp}(1\;\mathrm{Hz})]$, $\mathrm{real} [\mathrm{amp}(2\;\mathrm{Hz})]$), were compared to the 95\% confidence intervals (CI) generated by the distribution of permuted activations. Neurons whose $\mathrm{real} [\mathrm{amp}(1\;\mathrm{Hz})]$ and $\mathrm{real} [\mathrm{amp}(2\;\mathrm{Hz})]$ values exceeded this threshold were classified as \textit{significant MLP neurons} (see \ref{defi1}), indicating their involvement in syntactic processing with statistical robustness against random noise.

\begin{defi}[Significant MLP Neurons]
    \label{defi1}
    For a fixed frequency $f$, a neuron is a significant MLP neuron, if and only if its FFT result satisfies
    \begin{equation}
        \mathrm{real} [\mathrm{amp}(f)] \notin \text{95\% CI of permuted distribution}.
    \end{equation}
    The set containing all the significant MLP neurons in terms of frequency $f$ is denoted as $\mathbb{S}_f$.
\end{defi}

Since the \textit{significant MLP neurons} are distributed almost uniformly across all layers, detecting the specific neurons that contribute to sentence and phrase processing requires a more objective and systematic method. We then applied z-scores to the FFT amplitudes at \(1\;\mathrm{Hz}\) and \(2\;\mathrm{Hz}\) in both the experimental and control groups for all \textit{significant MLP neurons} across layers. The z-score deviation $z_f(n)$ between the experimental and control groups was then calculated for each neuron.This deviation helps minimize semantic confounds by isolating frequency-specific syntactic effects. Sentence and phrase MLP neurons were defined as those whose z-scores deviated by more than two standard deviations from the mean, at \(1\;\mathrm{Hz}\) and \(2\;\mathrm{Hz}\), respectively (see \ref{defi2}). 

\begin{defi}[Sentence MLP Neurons and Phrase MLP Neurons]
\label{defi2}
    A neuron $n$ is defined as a sentence/phrase MLP neuron if it satisfies 
    \begin{equation}
        n\in\mathbb{S}_f,\quad z_f(n)\geq \mu_{z_f}+2\sigma_{z_f},
    \end{equation}
    where $z_f(n)$ denotes the $z$-score deviation of the FFT amplitude between experimental and control groups for neuron $n$ at frequency $f$, $\mu_{z_f}$ denotes the mean $z$-score across all neurons for the frequency $f$, $\sigma_{z_f}$ denotes the standard deviation of $z$-scores across all neurons for the frequency $f$, and the frequency $f$ is specified as $1\;\mathrm{Hz}$ and $2\;\mathrm{Hz}$ for sentence and phrase MLP neuron respectively.
\end{defi}

Following this, we identified and analysed sentence and phrase MLP neurons across layers and LLMs, with full details provided in Section \ref{LLM_experiment}. We also conducted bilingual experiments to assess the ability of different LLMs to perceive syntactic structures across Chinese and English (see Appendix \ref{multilingual}).

\subsection{Syntactic structure probe in the human brain}
\label{process_human}
We recorded sEEG from 26 native Chinese speakers while they listened to two Chinese auditory corpora. In the \textbf{sentence} corpus, nine four‑syllable sentences were concatenated per trial; in the \textbf{phrase} corpus, eighteen two‑syllable phrases were concatenated. Each corpus comprised 40 trials per subject. Syllables were 250 ms long, and signals were sampled at \(512\;\mathrm{Hz}\) (\(512\;\mathrm{Hz}\) for one participant). To minimize onset‑related responses, analyses used only the final 32 syllables of each trial.

To analyze the sEEG data, we employed inter-trial phase coherence (ITPC), a frequency-domain method relatively resistant to noise that quantifies the consistency of phase relationships in oscillatory brain activity across multiple trials \citep{cohen2014analyzing}. sEEG Electrode localization was performed similarly to previous studies \citep{xu2023two, wang2024neural}; all electrodes were mapped to brain regions defined by the Automated Anatomical Labeling (AAL) system. We then grouped certain AAL regions to form 12 brain regions of interest (ROIs) (details in Appendix \ref{region_label}). Subsequent experiments were conducted based on brain ROIs.

As previously outlined, the proposed HFTP approach is designed to be applicable to both LLMs and human brain data. For the human brain analysis, we employed the same permutation testing procedure on the time-domain sEEG data that captured cortical activity during listening to Chinese corpora. Specifically, ITPC results were randomized 1000 times for each channel in each subject. The original frequency bins, $\mathrm{real} [\mathrm{amp}(1\;\mathrm{Hz})]$ and $\mathrm{real} [\mathrm{amp}(2\;\mathrm{Hz})]$, were then assessed to determine whether they fell within the 95\% confidence interval of the permuted ITPC distribution (see \ref{defi3}).

\begin{defi}[Sentence channels and Phrase channels]
\label{defi3}
    A channel $c$ is defined as a sentence/phrase channel if its ITPC result satisfies
    \begin{equation}
        \mathrm{real}[\mathrm{amp}(f)]\notin \text{95\% CI of permuted ITPC},
    \end{equation}
    where $f=1\;\mathrm{Hz}$ for sentence channel and $f=2\;\mathrm{Hz}$ for phrase channel.
\end{defi}

Using this probe, we identified and analyzed the distribution of sentence and phrase channels across various brain ROIs, with full details provided in Section \ref{brain_experiments}.

\subsection{Alignment of syntactic structure representations of LLMs with the human brain}
\label{alignment_}

To explore syntactic structure representation alignment between LLMs and the human brain, we compared their frequency-domain representations using the same \textbf{sentence} and \textbf{phrase} corpora. For each computational module (a MLP neuron in LLMs or an sEEG electrode channel in the human brain), we extracted amplitude in the frequency spectrum as the feature. This approach creates a multi-dimensional space based on frequency-domain features, where each syntactic structure corresponds to a specific point in this space (see Figure \ref{searchlight}). We then computed the distances between these points for different syntactic structures within the same computational module using cosine similarity. Through pairwise comparisons, we constructed Structure Representational Dissimilarity Matrices (SRDMs) for each computational module, which are similar to Representational Dissimilarity Matrices (RDMs) but specifically capture the representations of syntactic structures \citep{cichy2014comparison, khaligh2014deep}. We then applied Representational Similarity Analysis (RSA) to enable cross-modal comparisons between LLMs and brain data, correlating the representations in both systems \citep{kriegeskorteRepresentationalSimilarityAnalysis2008}. This approach quantified alignment and used statistical tests to detect significant overlaps. We introduced two key measures: model-brain similarity $S(m,b)$ and model-region similarity $S(m,b_r)$, to evaluate alignment globally and in specific brain ROI, and used the contribution ratio ($CR_r$) to assess the impact of each region on the alignment. For more details on the alignment pipeline, see Appendix \ref{alignment_pipeline}. The comprehensive discussion of the alignment results can be found in Section \ref{align_experiment}.

\section{Experiments}

We used HFTP to assess structural processing in the human brain and LLMs, and aligned their frequency-domain representations to evaluate their similarity.

\subsection{MLP neurons represent sentences and phrases in LLMs}
\label{LLM_experiment}
\begin{figure*}[ht]
\begin{center}
% \captionsetup{aboveskip=-2pt}
\centerline{\includegraphics[width=\linewidth]{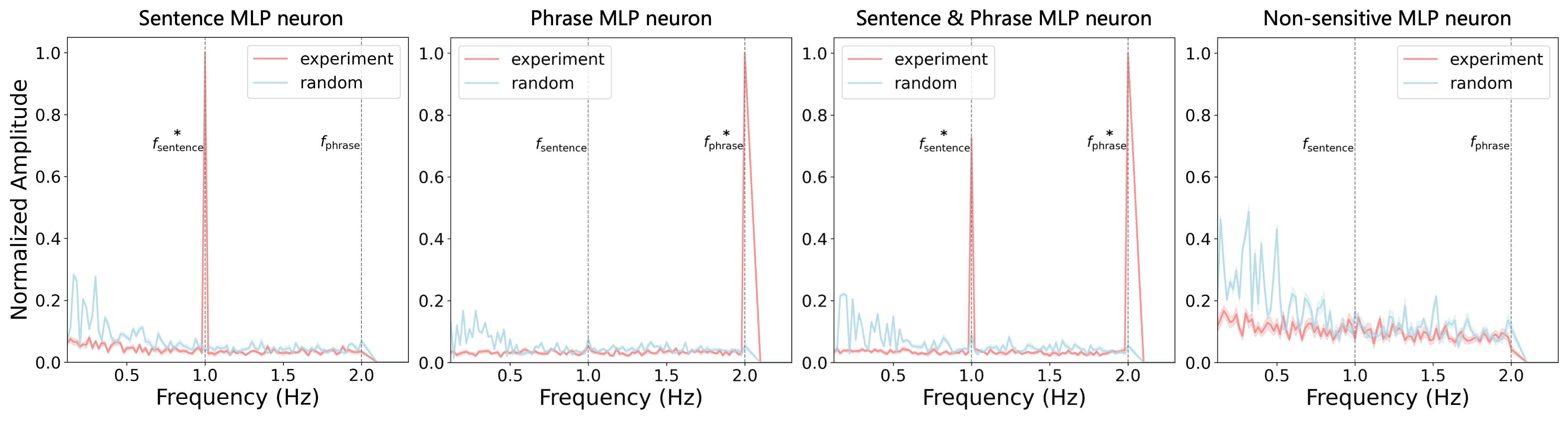}}
\caption{Hierarchical frequency patterns of MLP neurons selectively represent sentence features, phrase features, shared features of both and non-sensitive feature (from left to the right). Here, ``experiment" denotes the original corpus, while ``random" indicates the randomized version. Shaded bands show \(\pm 1\;\mathrm{s.e.m.}\) computed across 10 shuffled-input activation partitions. Significant peaks (*$p < 0.05$, FDR corrected) indicate amplitudes stronger than neighboring frequencies within \(\pm 0.5\;\mathrm{Hz}\). "Normalized Amplitude" represents the curves and bands scaled to a range of 0 to 1.}
    \label{fft-example}
\end{center}
\vskip -0.3in
\end{figure*}

Using the HFTP method, we identified neurons in all six models that selectively represent sentences (sentence neurons), phrases (phrase neurons), and neurons that simultaneously represent both (sentence \& phrase neurons). In the examples shown, we highlight MLP neurons which display distinct hierarchical frequency patterns. Figure \ref{fft-example} demonstrates four patterns: a significant peak at the sentence frequency ($f_{\mathrm{sentence}}=1\;\mathrm{Hz}$), a significant peak at the phrase frequency ($f_{\mathrm{phrase}}=2\;\mathrm{Hz}$), dual peaks at both $f_{\mathrm{sentence}}$ and $f_{\mathrm{phrase}}$, and no significant peaks. Frequencies beyond \(2\;\mathrm{Hz}\) have been artificially set to zero for smoothness in the representation.

\begin{figure*}[ht]
% \captionsetup{belowskip=-6pt} % Reduce space between caption and text
\centering
    % First row: Three images
    \begin{subfigure}{0.3\textwidth}
        \centering
        \includegraphics[width=\linewidth]{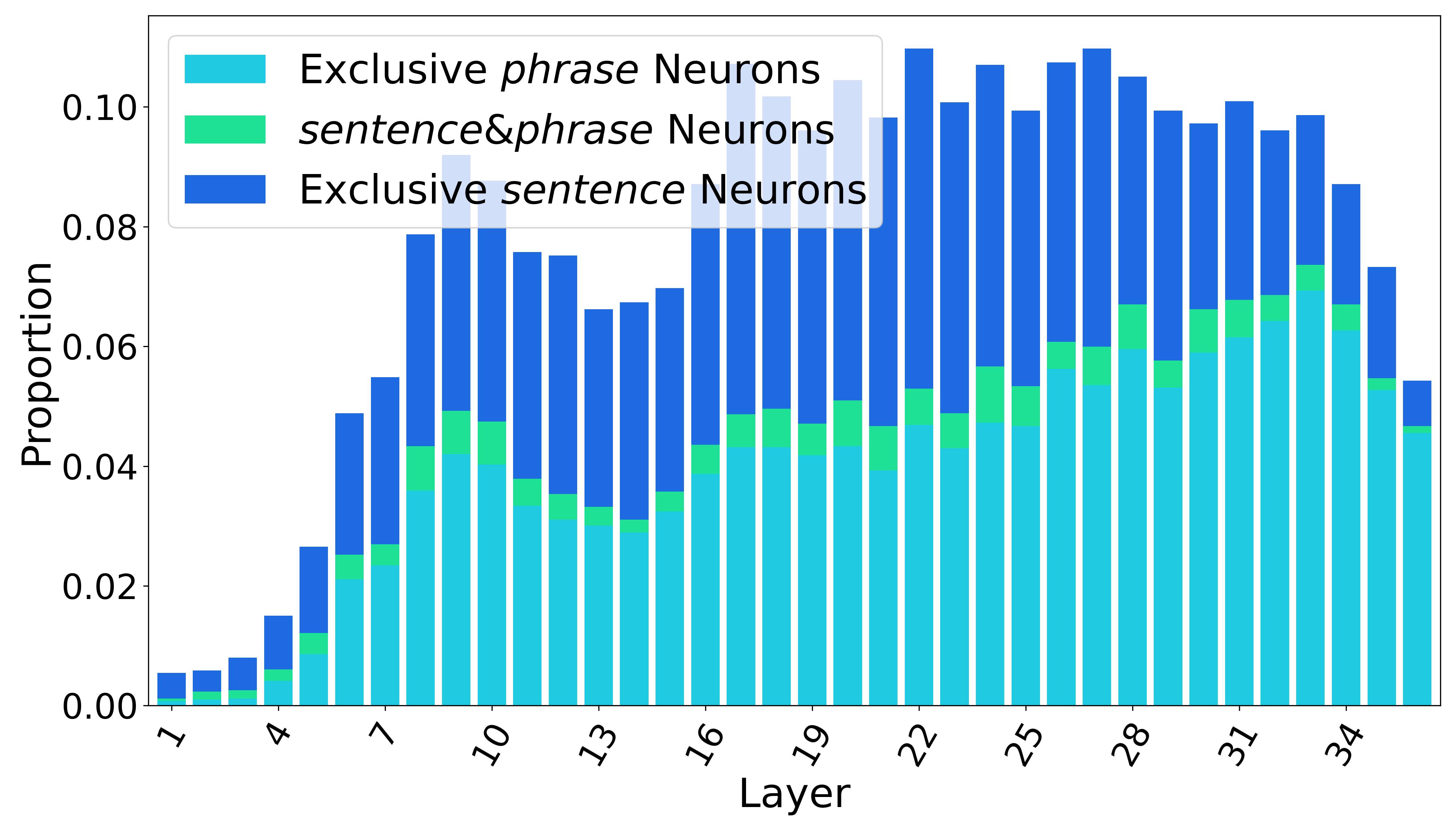}
        \subcaption{GPT-2}
        \label{gpt2-ssvoc}
    \end{subfigure}
    \hspace{1em}
    \begin{subfigure}{0.3\textwidth}
        \centering
        \includegraphics[width=\linewidth]{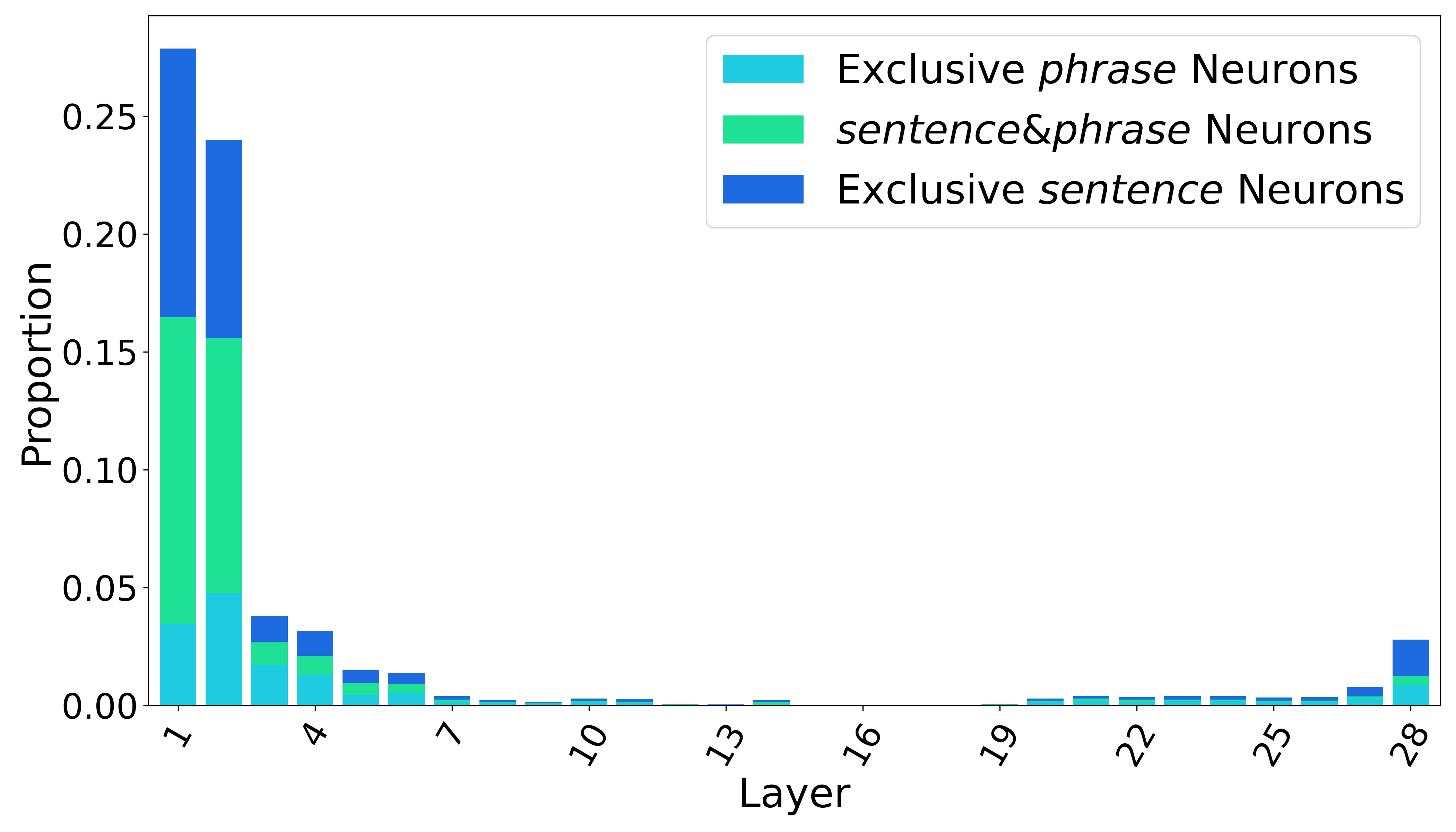}
        \subcaption{Gemma}
        \label{gemma-ssvoc}
    \end{subfigure}
    \hspace{1em}
    \begin{subfigure}{0.3\textwidth}
        \centering
        \includegraphics[width=\linewidth]{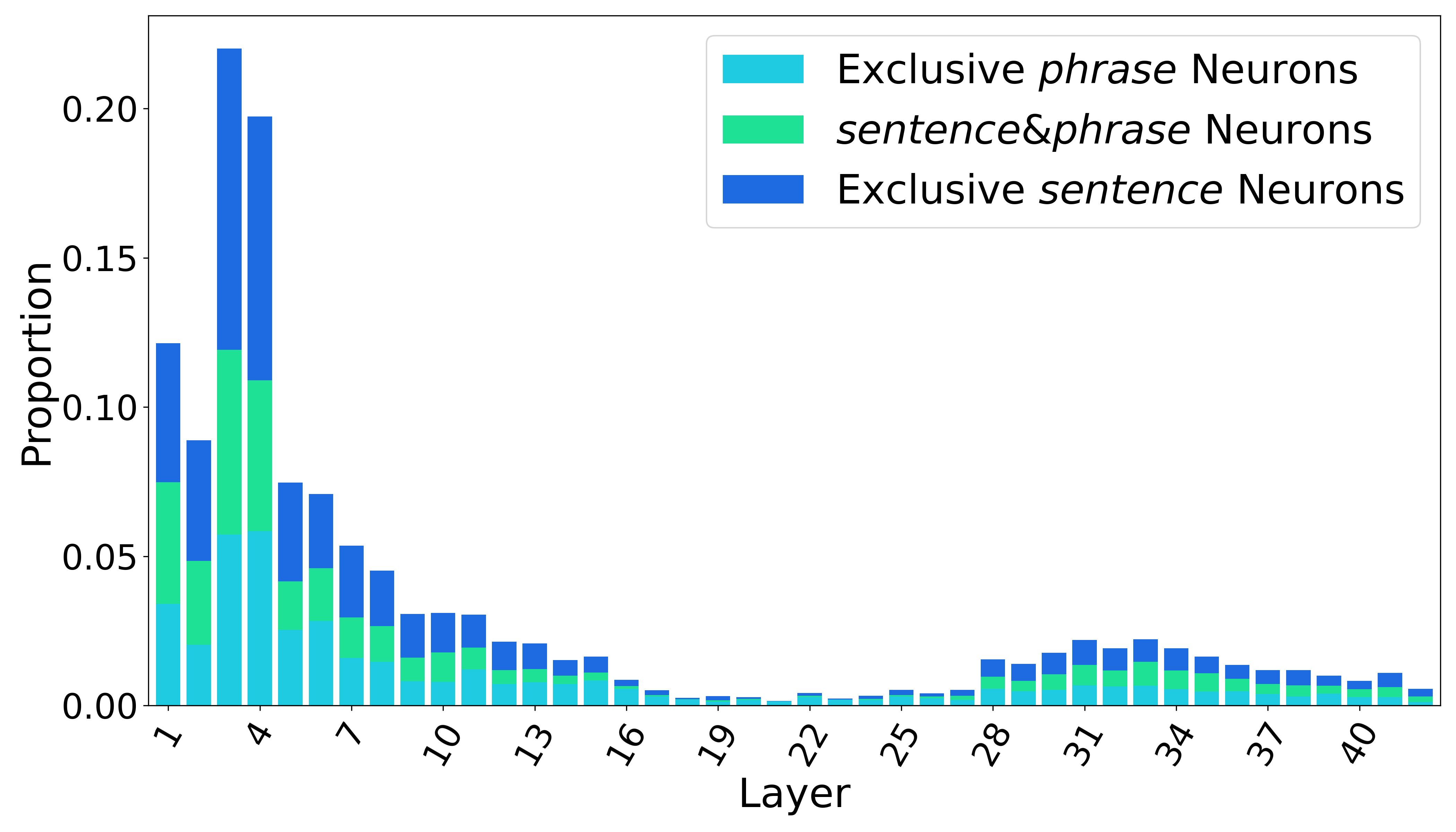}
        \subcaption{Gemma 2}
        \label{gemma2-ssvoc}
    \end{subfigure}

    \vspace{1em} % Optional: Adjust vertical space between rows

    % Second row: Three images
     \begin{subfigure}{0.3\textwidth}
        \centering
        \includegraphics[width=\linewidth]{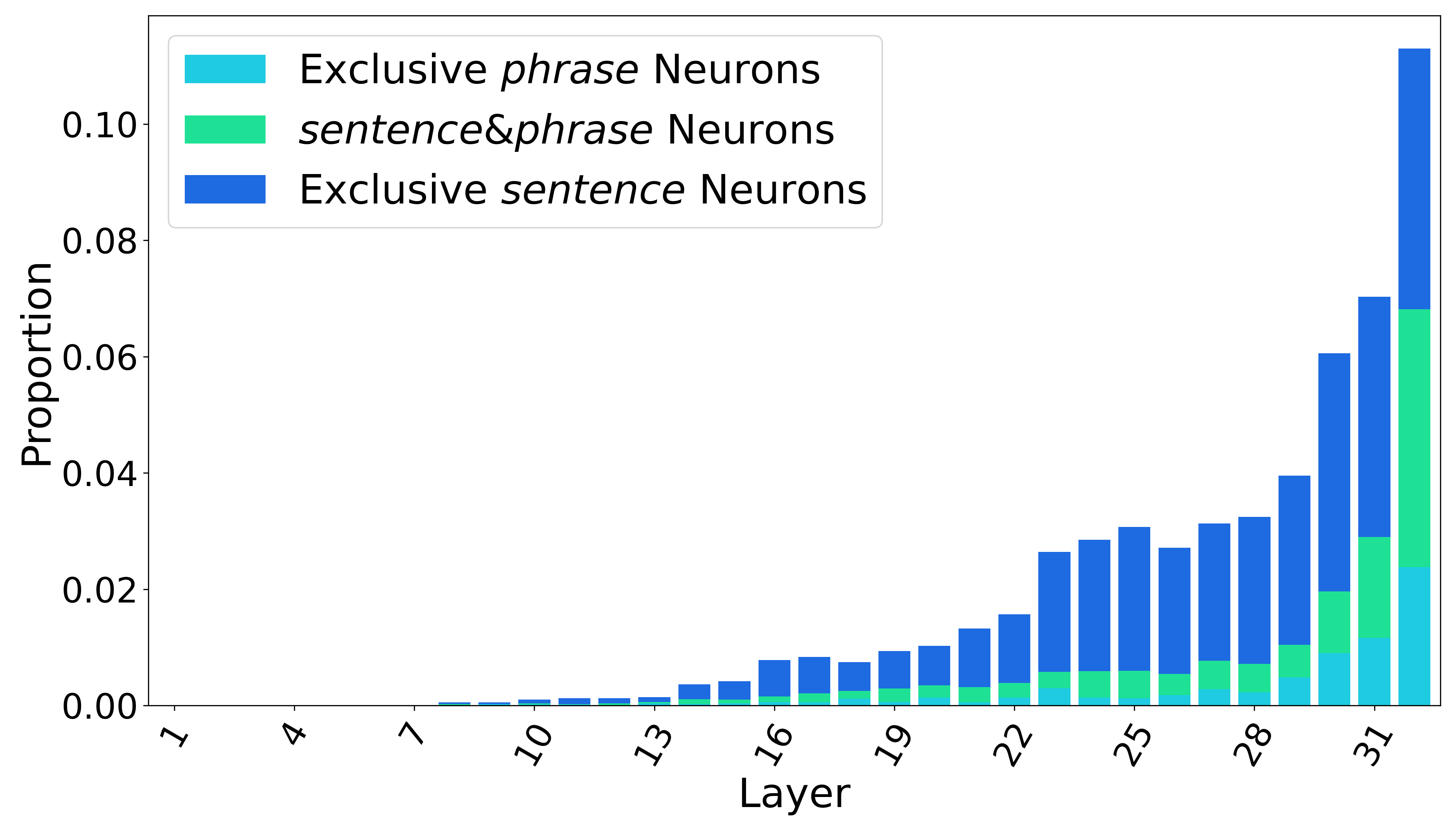}
        \subcaption{Llama 2}
        \label{llama2-ssvoc}
    \end{subfigure}
    \hspace{1em}
    \begin{subfigure}{0.3\textwidth}
        \centering
        \includegraphics[width=\linewidth]{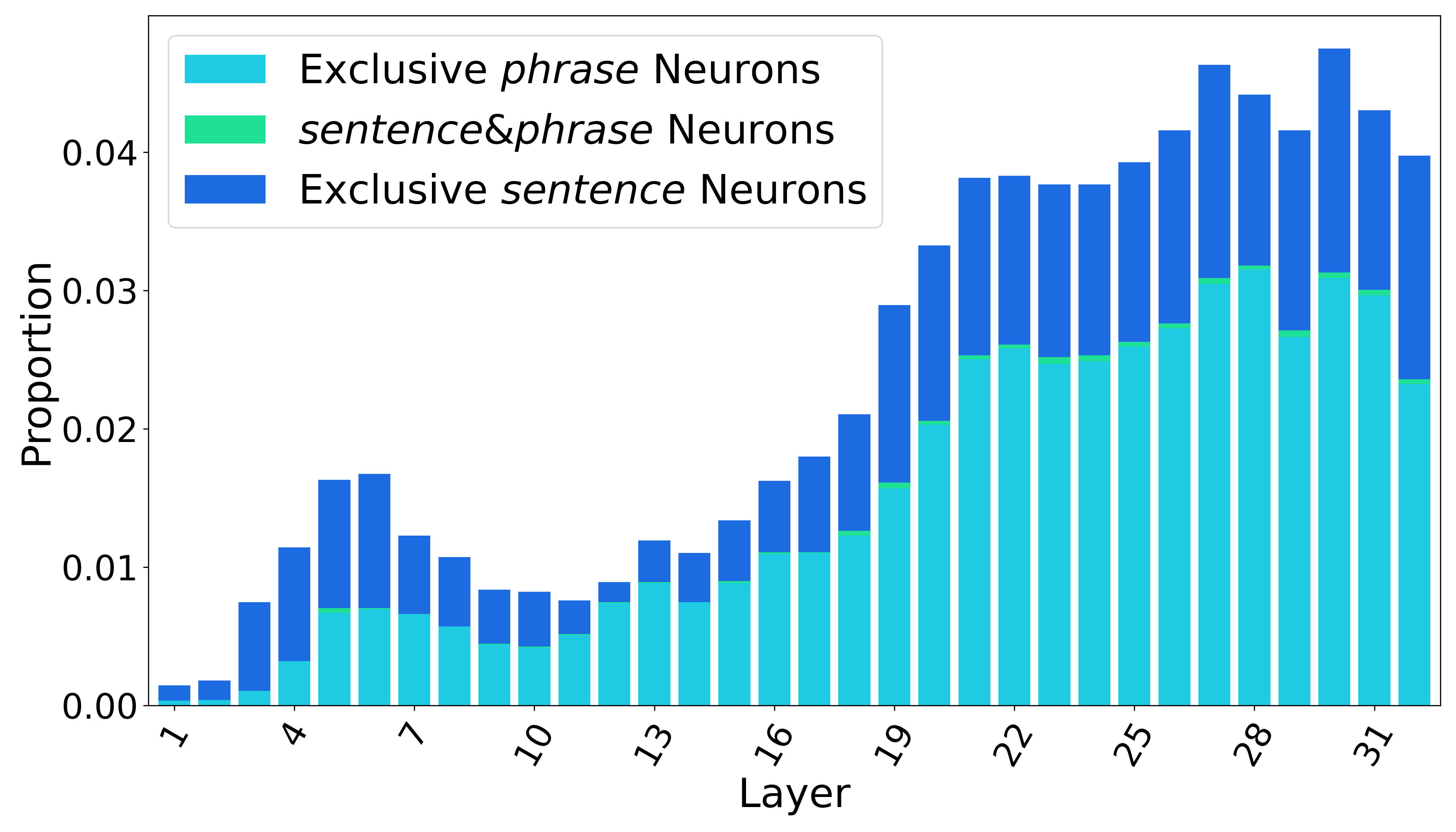}
        \subcaption{Llama 3.1}
        \label{llama3-ssvoc}
    \end{subfigure}
    \hspace{1em}
    \begin{subfigure}{0.3\textwidth}
        \centering
        \includegraphics[width=\linewidth]{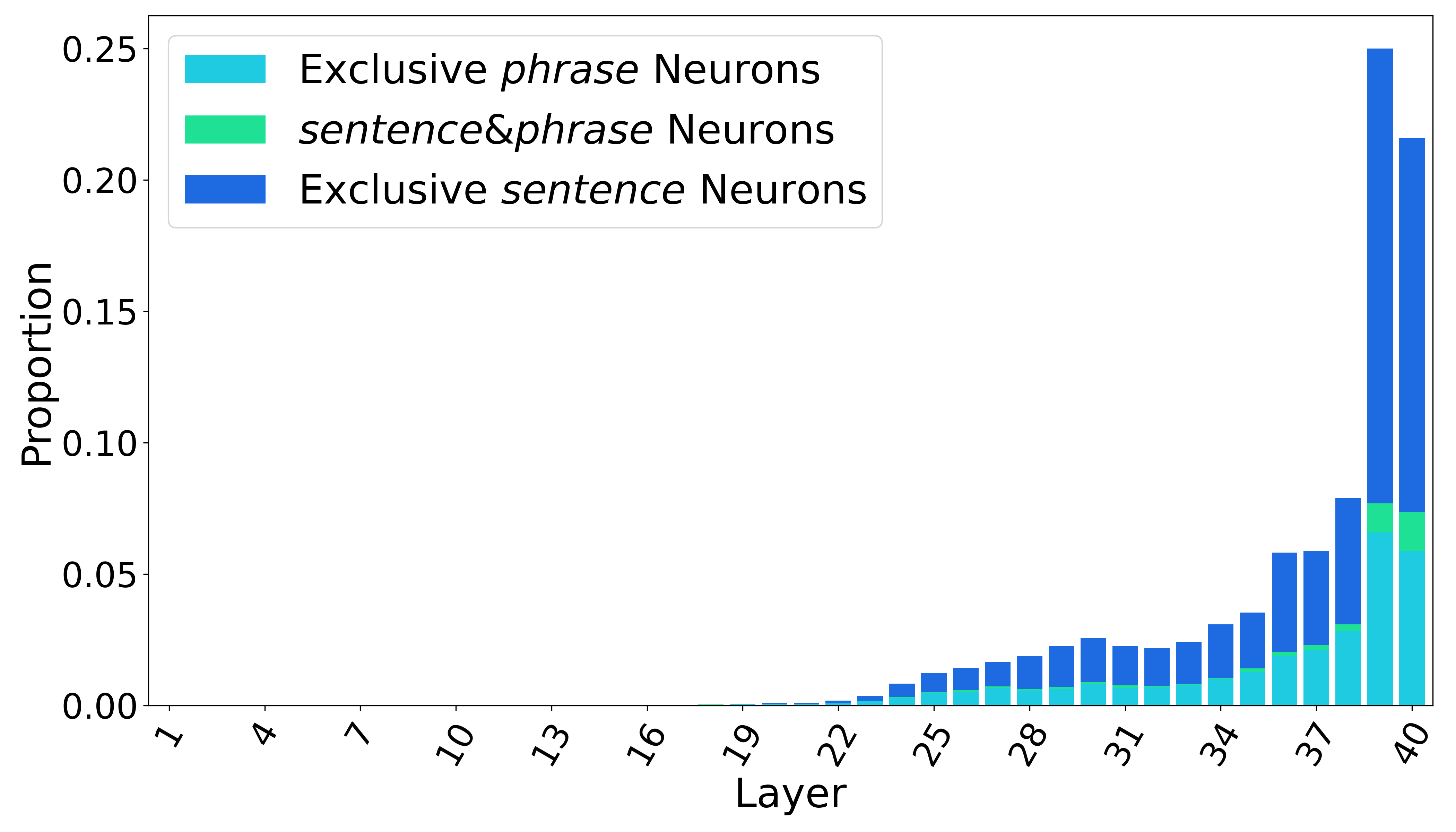}
        \subcaption{GLM-4}
        \label{glm-ssvoc}
    \end{subfigure}

    \caption{Statistics of exclusive sentence/phrase MLP neurons and sentence \& phrase MLP neurons in each layer across six LLMs.}
    \label{all-ssvoc}
    \vskip -0.15in
\end{figure*}

\begin{figure*}[ht]
\begin{center}
% \captionsetup{aboveskip=-2pt}
% \captionsetup{belowskip=-6pt} % Reduce space between caption and text
\centerline{\includegraphics[width=\linewidth]{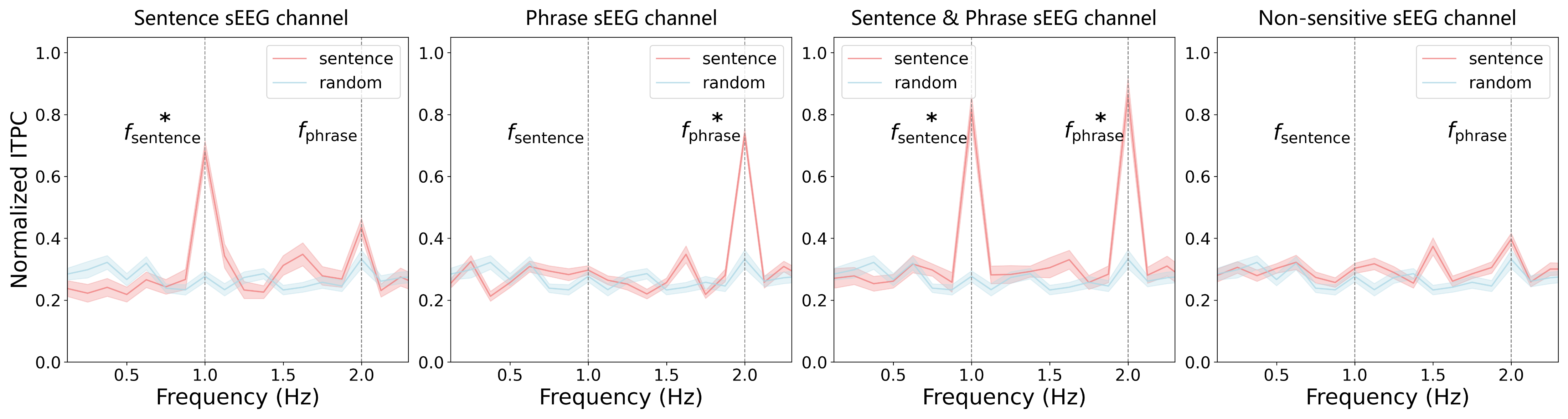}}
 \caption{Hierarchical frequency patterns of sEEG channels selectively represent sentence features, phrase features, shared features of both  and non-sensitive feature (from left to the right). Here, ``setence" denotes the original \textbf{sentence} corpus, while ``random" indicates the randomized version. Shaded bands show \(\pm 1\;\mathrm{s.e.m.}\) computed across channels. Significant peaks (*$p < 0.05$, FDR corrected) indicate amplitudes stronger than neighboring frequencies within \(\pm 0.5\;\mathrm{Hz}\). "Normalized ITPC" represents the curves and bands scaled to a range of 0 to 1.}
    \label{itpc}
\end{center}
\vskip -0.2in
\end{figure*}

For the six LLMs tested, we identified their sentence and phrase neurons using the HFTP method. Figure \ref{all-ssvoc} shows the distribution of exclusive sentence/phrase neurons and those representing both across different layers, based on experiments using the Chinese syntactic corpus. All the models contain neurons dedicated to capturing sentences and phrases, demonstrating their ability to encode the syntactic hierarchies of human language. However, distinct distribution patterns suggest varied syntactic processing strategies: Llama and GLM primarily process syntactic information in later layers, reflecting a more integrated approach. GPT, on the other hand, exhibits higher concentrations of sentence and phrase neurons in its middle layers, suggesting a balanced intermediate strategy. In contrast, Gemma demonstrates a distinct preference for dense concentrations of neurons in the early layers, indicating an emphasis on initial-stage syntactic processing.

A comparative analysis reveals a significant decrease in the maximum proportions of all syntactic neurons in the Llama and Gemma models, dropping from 11\% in Llama 2 to 4.5\% in Llama 3.1, and from 27\% in Gemma to 22\% in Gemma 2. As Llama 3.1 and Gemma 2 are updated versions of Llama 2 and Gemma, respectively, this trend suggests a potential shift in computational resource allocation. To enhance performance on complex tasks, Llama 3.1 and Gemma 2 may reduce their specialized processing of syntactic structures (sentences and phrases), reallocating neurons to support these advanced capabilities.

Additionally, a consistent covariant trend between sentence and phrase neurons across layers was observed for all six models, with high statistical correlations, including GPT-2 (r = 0.754), Gemma (r = 0.994), Gemma 2 (r = 0.994), Llama 2 (r = 0.912), and Llama 3.1 (r = 0.886), GLM (r = 0.993). This suggests that LLMs share similar underlying mechanisms for sentence and phrase processing.

% Note use of \abovespace and \belowspace to get reasonable spacing
% above and below tabular lines.

\subsection{Sentences and phrases representations in the human brain}
\label{brain_experiments}
Using the HFTP approach, we identified neuron populations in the human brain that selectively represent sentences and phrases. Each sEEG channel captures collective responses from nearby neuron populations, providing high spatio-temporal resolution of neural activity. This allows us to assess sentence and phrase selectivity precisely. As shown in Figure \ref{itpc}, we found channels representing sentences and phrases, as well as channels with shared representations, while some channels did not represent either. These findings align with those observed in LLMs, demonstrating that HFTP effectively investigates the internal representations of syntactic structures in both systems.

\begin{figure}[ht]
    \centering
    % \captionsetup{belowskip=-6pt} % Reduce space between caption and text
    \begin{subfigure}[b]{0.475\linewidth}
        \centering
        \raisebox{0.8cm}{ % Raise the left image by 2cm to vertically align with the right image
            \includegraphics[width=\linewidth]{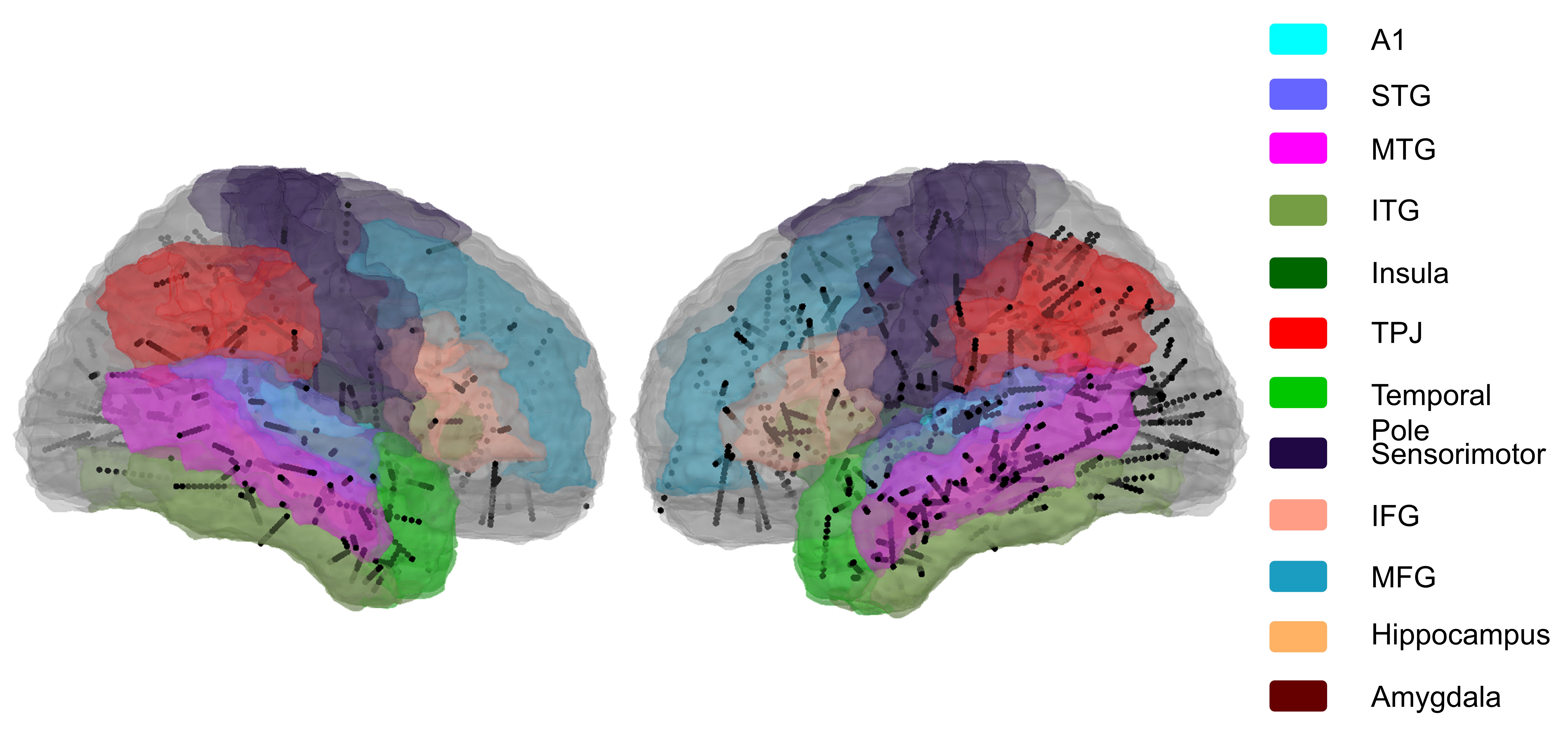}
        }
        \caption{sEEG channel locations and Brain ROIs}
    \end{subfigure}
    \begin{subfigure}[b]{0.475\linewidth}
        \centering
        \includegraphics[width=\linewidth]{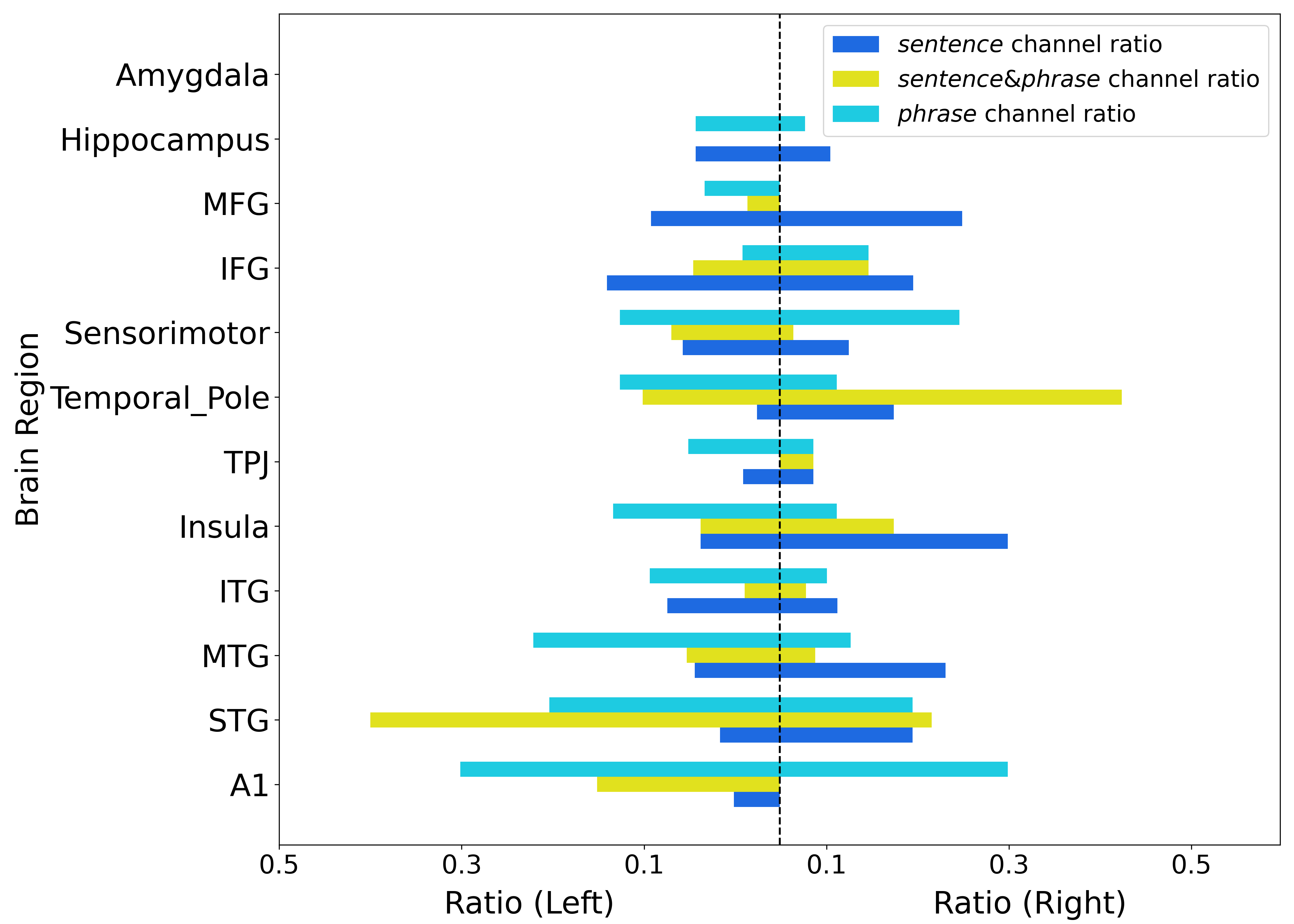}
        \caption{Significant sEEG channel distribution}
    \end{subfigure}
    \caption{(a) Brain ROIs of the left and right hemispheres used in this study. The black electrodes represent the sEEG channel locations across all participants. (b) Distribution of significant exclusive sentence/phrase and sentence \& phrase channels (\textbf{sentence} corpus) in different brain ROIs.}
    \label{distri_region}
    % \vskip -0.15in
\end{figure}

\begin{table}[ht]
\centering
% \caption{Averaged top 100 Spearman correlation coefficients between SRDMs of sEEG channels and those of MLP neurons under the \textbf{sentence} corpus condition, separated by left (L) and right (R) hemispheres. Note that '/' denotes cases where the model lacks channels corresponding to the brain ROI in that hemisphere within the top 100 Spearman-ranked channels. The values in the rows corresponding to brain ROIs represent the model-region similarity $S(m,b_r)$.} 
\caption{Averaged top-100 Spearman correlations between sEEG-channel SRDMs and MLP-neuron SRDMs under the \textbf{sentence} corpus, separated by left (L) and right (R) hemispheres. ‘/’ indicates that no channels from that hemisphere appear in the top-100 set. The first data row reports the overall model–brain similarity \(S(m,b)\); each subsequent brain ROI row reports the model–region similarity \(S(m,b_r)\). Bold in the \(S(m,b)\) row flags the model with the highest overall alignment, while bold entries within each brain ROI row mark the three regions most strongly correlated for that model.}
\label{sim_sentence}
% \vskip 0.15in
\begin{small}
\begin{tabularx}{\textwidth}{@{}c*{13}{>{\centering\arraybackslash}X}@{}}
        \toprule
        &  \multicolumn{2}{c}{GPT-2} &  \multicolumn{2}{c}{Gemma} &  \multicolumn{2}{c}{Gemma 2} &  \multicolumn{2}{c}{Llama 2} &  \multicolumn{2}{c}{Llama 3.1} &  \multicolumn{2}{c}{GLM-4} \\ 
        \cmidrule(lr){2-13}
        &  L &  R &  L &  R &  L &  R &  L &  R &  L &  R &  L &  R \\ 
        \midrule
        $S(m,b)$ & \textbf{0.654} & 0.442 & 0.582 & 0.411 
&  0.644 &  0.450 & 0.645 & 0.439 & 0.514 & 0.405 & 0.630 & 0.445 
\\ 
        \midrule
        A1 & \textbf{0.683} & 0.423 & \textbf{0.642} & 0.358 
&  \textbf{0.702} &  0.333 & 0.649 & 0.547 & 0.514 & 0.403 & \textbf{0.664} & 0.374 
\\ 
        STG & 0.667 & 0.422 & \textbf{0.593} & 0.386 
&  0.654 &  0.410 & \textbf{0.672} & 0.453 & 0.507 & 0.392 & \textbf{0.647} & 0.412 
\\ 
        MTG & 0.674 & 0.392 & 0.584 & 0.383 
&  \textbf{0.659} &  0.411 & \textbf{0.67}4 & 0.409 & \textbf{0.521} & 0.408 & 0.645 & 0.397 
\\ 
        ITG & 0.637 & 0.444 & 0.578 & 0.406 
&  0.631 &  0.448 & 0.629 & 0.426 & 0.509 & 0.401 & 0.615 & 0.439 
\\ 
        Insula & 0.624 & 0.460 & 0.551 & 0.425 
&  0.600 &  0.476 & 0.630 & 0.446 & \textbf{0.518} & 0.422 & 0.604 & 0.475 
\\ 
        TPJ & 0.610 & 0.452 & 0.566 & 0.373 
&  0.641 &  0.410 & 0.619 & 0.400 & \textbf{0.518} & 0.408 & 0.606 & 0.438 
\\ 
        Temporal Pole & 0.648 & 0.473 & 0.556 & 0.470 
&  0.643 &  0.558 & 0.610 & 0.469 & 0.494 & 0.448 & 0.616 & 0.483 
\\ 
        Sensorimotor & 0.637 & 0.462 & 0.567 & 0.426 
&  0.622 &  0.448 & 0.624 & 0.446 & 0.505 & 0.396 & 0.617 & 0.463 
\\ 
        IFG & \textbf{0.694} & 0.463 & \textbf{0.603} & 0.466 
&  \textbf{0.670} &  0.496 & \textbf{0.665} & 0.491 & 0.513 & 0.410 & \textbf{0.646} & 0.490 
\\ 
        MFG & 0.615 & 0.436 & 0.557 & 0.401 
&  0.585 &  0.489 & 0.597 & 0.367 & 0.510 & 0.397 & 0.588 & 0.473 
\\ 
        Hippocampus & \textbf{0.698} & 0.405 & 0.553 & 0.408 
&  0.626 &  0.428 & 0.657 & 0.413 & 0.534 & 0.390 & 0.613 & 0.434 
\\ 
        Amygdala & /& 0.489 & 0.566 & 0.454 
&  /&  0.472 & /& 0.558 & 0.496 & 0.377 & /& 0.508 
\\ 
        \bottomrule
    \end{tabularx}
    \end{small}
    \vskip -0.2in
\end{table}

Similar to our analysis of neuron types in LLM layers, we calculated the proportions of sentence and phrase channels within each brain ROI. As shown in Figure \ref{distri_region}, phrase channels decrease from lower layers (A1) to higher layers (e.g., IFG), while sentence channels show the opposite trend, increasing at higher brain layers. This pattern aligns with earlier MEG studies \cite{sheng2019cortical}, supporting distinct processing mechanisms for sentences and phrases. Correlations between sentence and phrase channels across brain ROIs in both hemispheres revealed no significant relationship (left: r = -0.169, p = 0.870; right: r = -0.197, p = 0.539), suggesting that sentence and phrase processing operate independently. This contrasts with the behavior of LLMs, implying that while the human brain segregates sentence and phrase processing across different regions, LLMs integrate both syntactic levels within the same model layers. This highlights that the layered representations of LLMs may not align directly with the distinct processing roles observed in brain ROIs.

\subsection{Alignment of syntactic structure representations between LLMs and the human brain} 
\label{align_experiment}

As the layered representations of LLMs do not correspond directly to the distinct processing functions of different brain ROIs, we sought to investigate whether overall syntactic structure representations in LLMs are comparable to those in the human brain, both globally and across individual brain ROIs. To accomplish this, we employed Searchlight representational alignment at both sentence and phrase levels (see Section \ref{alignment_pipeline} for detailed procedures).

The alignment results across both \textbf{sentence} and phrase corpus (Tables \ref{sim_sentence} and \ref{similarity_phrase}) revealed consistent patterns of model-brain correspondence. As shown in both tables, we observed that the similarity between LLMs and the left hemisphere is notably higher than that with the right hemisphere across both linguistic granularities. One-way ANOVAs at the brain level confirmed that models differed significantly in explaining neural responses for both sentence-level ($F = 59.74$, $\eta^2 = 0.027$, $p < 0.001$) and phrase-level processing ($F = 12.10$, $\eta^2 = 0.018$, $p < 0.001$). Region-wise ANOVAs with FDR correction further showed a strongly left-lateralized effect at both corpus, concentrated in core language cortex (e.g., STG, MTG, IFG).

\begin{table}[ht]
\begin{small}
\centering
% \caption{Averaged top 100 Spearman correlation coefficients between SRDMs of sEEG channels and those of MLP neurons under the \textbf{phrase} corpus condition, separated by left (L) and right (R) hemispheres. Note that '/' denotes cases where the model lacks channels corresponding to the brain ROIs in that hemisphere within the top 100 Spearman-ranked channels. The values in the rows corresponding to brain ROIs represent the model-region similarity $S({m,b_r})$.}
\caption{Averaged top-100 Spearman correlations between sEEG-channel SRDMs and MLP-neuron SRDMs under the \textbf{phrase} corpus, separated by left (L) and right (R) hemispheres. ‘/’ indicates that no channels from that hemisphere appear in the top-100 set. The first data row reports the overall model–brain similarity \(S(m,b)\); each subsequent brain ROI row reports the model–region similarity \(S(m,b_r)\). Bold in the \(S(m,b)\) row flags the model with the highest overall alignment, while bold entries within each brain ROI row mark the three regions most strongly correlated for that model.}
\label{similarity_phrase}
% \vskip 0.15in
\begin{tabularx}{\textwidth}{@{}c*{13}{>{\centering\arraybackslash}X}@{}}
    \toprule
    & \multicolumn{2}{c}{GPT-2} & \multicolumn{2}{c}{Gemma} & \multicolumn{2}{c}{Gemma 2} & \multicolumn{2}{c}{Llama 2} & \multicolumn{2}{c}{Llama 3.1} & \multicolumn{2}{c}{GLM-4} \\ 
    \cmidrule(lr){2-13}
    & L & R & L & R & L & R & L & R & L & R & L & R \\ 
    \midrule
    $S({m,b})$ & \textbf{0.654} & 0.441 & 0.575 & 0.416 
&  0.628 &  0.443 & 0.648 & 0.435 & 0.522 & 0.404 & 0.626 & 0.437 
\\
    \midrule
    A1 & \textbf{0.669} & 0.420 & \textbf{0.610} & 0.315 
&  \textbf{0.665} &  0.326 & 0.628 & 0.576 & \textbf{0.526} & 0.372 & 0.627 & 0.374 
\\ 
    STG & 0.665 & 0.415 & \textbf{0.591} & 0.388 
&  0.643 &  0.395 & \textbf{0.674} & 0.447 & \textbf{0.530} & 0.402 & \textbf{0.649} & 0.401 
\\ 
    MTG & 0.665 & 0.391 & 0.564 & 0.408 
&  \textbf{0.645} &  0.415 & \textbf{0.675} & 0.415 & 0.521 & 0.401 & \textbf{0.641} & 0.408 
\\ 
    ITG & 0.637 & 0.447 & 0.558 & 0.421 
&  0.611 &  0.440 & 0.632 & 0.421 & 0.518 & 0.398 & 0.621 & 0.427 
\\ 
    Insula & 0.627 & 0.468 & 0.556 & 0.433 
&  0.596 &  0.465 & 0.632 & 0.453 & 0.519 & 0.409 & 0.602 & 0.470 
\\ 
    TPJ & 0.616 & 0.442 & 0.563 & 0.405 
&  0.608 &  0.423 & 0.615 & 0.384 & \textbf{0.542} & 0.399 & 0.588 & 0.433 
\\ 
    Temporal Pole & \textbf{0.680} & 0.453 & 0.555 & 0.394 
&  0.626 &  0.501 & 0.623 & 0.456 & 0.505 & 0.450 & 0.583 & 0.442 
\\ 
    Sensorimotor & 0.645 & 0.475 & 0.565 & 0.443 
&  0.607 &  0.449 & 0.629 & 0.442 & 0.511 & 0.391 & 0.613 & 0.460 
\\ 
    IFG & \textbf{0.700} & 0.450 & \textbf{0.635} & 0.424 
&  \textbf{0.656} &  0.491 & \textbf{0.674} & 0.488 & 0.514 & 0.436 & \textbf{0.636} & 0.480 
\\ 
    MFG & 0.618 & 0.409 & 0.562 & 0.421 
&  0.583 &  0.500 & 0.601 & 0.365 & 0.505 & 0.379 & 0.577 & 0.475 
\\ 
    Hippocampus & 0.656 & 0.398 & 0.541 & 0.418 
&  0.602 &  0.432 & 0.640 & 0.398 & 0.516 & 0.406 & 0.566 & 0.434 
\\ 
    Amygdala & /& 0.483 & /& 0.433 
&  /&  0.475 & /& 0.552 & 0.468 & 0.386 & /& 0.492 
\\ 
    \bottomrule
\end{tabularx}
\end{small}
\vskip -0.1in
\end{table}

% Similarly, phrase-level experiments (see Table \ref{similarity_phrase}) yielded comparable findings, reinforcing the robustness of the HFTP approach and suggesting its potential as a valuable tool for future studies of model-brain alignment. As we can see from Table \ref{similarity_phrase}, the alignment results highlight the effectiveness of the HFTP in capturing phase-level syntactic structure representations. At the brain level, a one-way ANOVA also revealed significant between-model differences ($F = 12.10$, $\eta^2 = 0.018$, $p < 0.001$), and region-wise ANOVAs (FDR-corrected) indicated that significant effects were primarily confined to left language-dominant areas (e.g., MTG, ITG, STG). Similar to the findings for sentence-level processing, the similarity with the left hemisphere is generally higher than with the right hemisphere. GPT-2 still exhibits a strong average correlation ($S$=0.654, L). Gemma 2 witness a improved correlation($S$=0.628, L) compared than Gemma ($S$=0.575, L). In contrast, Llama 3.1 shows a lower correlation ($S$=0.522, L) compared to Llama 2 ($S$=0.648, L), suggesting that improvements in model architecture do not necessarily result in better alignment with human brain activity. Additionally, other LLMs also demonstrate relatively high $S({m,b_r})$ values in these key structure processing brain ROIs as in Table \ref{sim_sentence}.

Examining individual model performance, GPT-2 exhibited the highest average correlation with human brain activity across both sentence ($S$ = 0.654, L) and phrase levels ($S$ = 0.654, L). Gemma 2 ($S$ = 0.644, L for sentences; $S$ = 0.628, L for phrases) consistently outperformed Gemma ($S$ = 0.582, L for sentences; $S$ = 0.575, L for phrases) at both corpora, attributed to architectural improvements \citep{team2024gemma}. Most notably, Llama 3.1 ($S$ = 0.514, L for sentences; $S$ = 0.522, L for phrases) showed consistently lower alignment than Llama 2 ($S$ = 0.645, L for sentences; $S$ = 0.648, L for phrases) across both processing levels. This counterintuitive pattern is explained at the pretraining level: Llama 3.1 was trained on a substantially larger corpus emphasizing code, reasoning, and multilingual text, which dilutes language-specific regularities. Additionally, extensive reliance on synthetic data for capability-targeted curation introduces distributional shifts away from naturalistic language statistics \citep{dubey2024llama,touvron2023llama}. These findings echo evidence that scaling alone fails to secure robust predicate–argument structure—especially for long-range, boundary-sensitive roles—revealing persistent human–model structural gaps \citep{cheng2024potential}.

Additionally, the key brain ROIs for each model, which are primarily located in the left hemisphere, highlighted regions that are critical for syntactic processing at both sentence and phrase levels. These regions include the left A1, STG, MTG, and IFG, with many of the LLMs exhibiting particularly strong correlations in these areas across both linguistic granularities, emphasizing their role in syntactic functions. These converging findings across sentence and phrase levels reinforce the robustness of the HFTP approach and suggest its potential as a valuable tool for future studies of model-brain alignment.

\section{Conclusion}
\label{conclusion}
This study advances syntactic processing by introducing the Hierarchical Frequency Tagging Probe (HFTP), a unified framework for dissecting neuron‑wise sentence and phrase representations in LLMs, population‑level patterns in the human brain, and generalizing seamlessly to naturalistic text. The results reveal that while LLMs, such as GPT-2, Gemma, Llama 2, and others, exhibit hierarchical syntactic processing and alignment with left-hemisphere brain activity, the mechanisms underlying their representations diverge significantly from those in human cortical regions. Notably, newer models like Gemma 2 demonstrate improved alignment, whereas others, such as Llama 3.1, show weaker human-model correlations despite enhanced task performance. These findings underscore the need to refine LLM architectures for more human-like syntactic processing and establish HFTP as a bridge between computational linguistics and cognitive neuroscience. Finally, the societal implications of this work are two-sided: positively, HFTP can support safer, more controllable models and inform non-invasive diagnostics via spectral markers; negatively, the same interpretability could be misused to optimise persuasive manipulation and, if linked with personal neural data, undermine privacy. 

\section{Limitation}
\label{limitation}
Although we applied HFTP to both Chinese and English corpora for LLMs, our sEEG data were collected in China and primarily involve Chinese stimuli. We have preliminary English responses from a single Chinese native participant, but deeper cross-linguistic analyses are pending; future work will recruit native English speakers to validate alignment on English and enable rigorous cross-linguistic tests.Additionally, we evaluated only a small set of model architectures and parameter scales, so the key universal mechanisms driving model–brain syntactic alignment remain unclear and warrant further investigation.

\bibliography{arxiv}
\bibliographystyle{unsrtnat}  % 因bst文件缺失
%%%%%%%%%%%%%%%%%%%%%%%%%%%%%%%%%%%%%%%%%%%%%%%%%%%%%%%%%%%%

%%%%%%%%%%%%%%%%%%%%%%%%%%%%%%%%%%%%%%%%%%%%%%%%%%%%%%%%%%%%

\newpage

\appendix

\section{Alignment pipeline for syntactic processing between LLMs and the human brain}
\label{alignment_pipeline}
This appendix we provide the detailed pipeline used to align the syntactic structure representations in LLMs with those in the human brain, focusing on detecting and comparing sentence/phrase representations across both systems.

\textbf{Data and Experimental Setup}
To maintain consistency between the LLM and human experiments, we used the same two corpora: a \textbf{sentence} corpus (four-syllable Chinese sequence) and a \textbf{phrase} corpus (two-syllable Chinese sequence). The word-order randomized version of each corpus was used as a control condition, as detailed in Section \ref{data_llm}. Each corpus included 40 trials and each trial contains 36 syllables. For SRDM calculation, the corpora were divided into six experimental conditions, each with 20 trials. Sentence/phrase representations of the last 32 syllables were extracted from both LLMs and human subjects to reduce the strong responses at word onset in both systems. The representations are then transformed into the frequency domain.

\textbf{Frequency-Domain Transformation and Similarity Metrics}
For the LLMs, neuron activations were transformed using FFT to capture the frequency components of structure processing across the six conditions. From this transformation, we calculated the cosine similarity between each pair of conditions, constructing an SRDM for each MLP neuron. Similarly, for the human brain, we calculated the ITPC to capture frequency-domain representations for each brain channel, producing a channel SRDM.

To assess the structure alignment between LLMs and the human brain, we computed the Spearman correlation $\rho$ between the SRDM of each LLM layer and the SRDM of each brain channel. We then grouped the model SRDMs at the layer level by averaging the cosine similarity across neurons for a fixed layer. The top 100 most relevant brain channels for each model layer were identified based on Spearman correlation, and the overlap of sentence/phrase channels in these top 100 channels was evaluated using a chi-square test. A significant overlap indicated alignment in structure processing between LLMs and brain ROIs.

\textbf{Model-Brain Similarity and Model-Region Similarity}
Two key metrics were defined to quantify the structure alignment between LLMs and the human brain. The first, model-brain similarity $S(m,b)$, represents the overall similarity of syntactic processing between an LLM $m$ and the human brain $b$. It is calculated as the average Spearman correlation between the SRDM of each LLM layer and the top 100 most relevant brain channels:

\begin{equation} S(m,b) =\frac{1}{M}\sum_{j=1}^M \frac{1}{100} \sum_{i\in{\mathrm{top}(j)}}^{100} \rho(L_j, C_i),\end{equation}

where $M$ is the number of layers of an LLM;  $L_j$ and $C_i$ denote the model layer and the brain channel with the indices $j$ and $i$ respectively;  $\mathrm{top}(j)$ means the indices of top 100 channels in terms of model layer $L_j$; and $\rho(L_j, C_i)$ denotes the Spearman correlation between the model SRDM at layer $L_j$ and the SRDM for brain channel $C_i$.

The second metric, model-region similarity $S(m,b_r)$, measures the alignment between LLMs and specific brain ROIs. This is calculated by averaging the Spearman correlation for the top 100 channels within a particular brain ROI:

\begin{equation}
    S({m,b_r})=\frac{1}{M}\sum_{j=1}^M \frac{1}{n(j,r)}\sum_{i\in\mathrm{top}(j)\cap\mathbb{C}_r}^{n(j,r)}\rho(L_j,C_i),
\end{equation}
where $\mathbb{C}_r$ denotes the indices of all channels belonging to the specific region $r$, and $n(j,r)$ means the total number of indices in $\mathrm{top}(j)\cap\mathbb{C}_r$, i.e. the number of channels belonging to region $r$ and at the same time within the top 100 channels in terms of model layer $L_j$.

\section{Predictive encoding control analysis}
\label{sec:predictive_encoding}

To provide an orthogonal test of model--brain alignment that does not depend on representational similarity analysis, we implemented a predictive encoding model that asks whether layer-wise features extracted from an LLM can \emph{predict} the frequency–domain sEEG responses of individual channels. All experimental settings (e.g. sentence/phrase corpora, block structure) were matched to those in the RDM--RSA pipeline (Section~\ref{alignment_pipeline}). We also adopt the same summary metrics, model--brain similarity $S(m,b)$ and model--region similarity $S(m,b_r)$, so that the predictive encoding results are directly comparable to the SRDM-RSA results reported above.

\textbf{Model features}
For each model layer $L_j$ we first selected significant neurons using the same procedure as in the SRDM-RSA pipeline. Within a given block, the time courses of the significant neurons were averaged to obtain one 32-sample sequence per layer and block (the last 32 syllables). We then applied FFT to this layer-average and retained the complex coefficients within 0.5--2\,Hz. Each retained coefficient was represented by its \emph{real part} and \emph{imaginary part} (the canonical cosine and sine components), which we treat as two features at that frequency. With two blocks per corpus, we formed the predictor matrix $X_j\in\mathbb{R}^{N\times 2}$ by stacking samples across blocks and frequencies, where $K$ denotes the number of frequency bins in 0.5--2\,Hz and $N=2K$ reflects the two blocks times $K$ frequencies. Thus each of the $N$ samples corresponds to one (block, frequency) pair and is described by a two-dimensional feature vector containing the real and imaginary parts of the aligned model coefficient.

\textbf{Brain responses}
For each sEEG channel $C_i$, we computed ITPC exactly as in the SRDM-RSA analysis, yielding a complex spectrum per block. We then aligned the frequency axes of the model and brain spectra by nearest-neighbor matching (model sampling 4\,Hz; sEEG 512\,Hz) within the 0.5--2\,Hz band. The target for predictive modeling was the band-limited spectral profile of stimulus-locked synchrony for channel $C_i$; specifically, we concatenated the ITPC amplitudes across the two blocks and across all $K$ frequencies to obtain an observation vector $y_i\in\mathbb{R}^{N}$. This vector summarizes the oscillatory response of channel $C_i$ within the syntactic band while preserving block-specific structure.

\textbf{Predictive model and cross-validation}
For each pair $(L_j,C_i)$ we fit a ridge regression with standardized predictors,
\begin{equation}
\hat{\boldsymbol{\beta}}
=\arg\min_{\boldsymbol{\beta}}
\;\bigl\|y_i-X_j\boldsymbol{\beta}\bigr\|_{2}^{2}
+\alpha\|\boldsymbol{\beta}\|_{2}^{2},
\end{equation}
using five random splits (train/test = 70/30). The ridge penalty $\alpha$ was chosen \emph{only on the training split} via inner ridge-CV over a logarithmic grid; the fitted model was then evaluated on the held-out test data. Predictive accuracy for a split was defined as Spearman correlation $\rho$ between the predicted and observed test targets. We averaged the split-wise correlations to obtain a predictive score
\begin{equation}
P(L_j,C_i)\;=\;\mathbb{E}_{\text{splits}}\bigl[\rho\bigl(\hat{y}_i^{\,\text{test}},\,y_i^{\text{test}}\bigr)\bigr],
\end{equation}
and repeated this procedure independently for left and right hemispheres and for both corpora. Note that feature selection is performed exclusively on the model side; the sEEG data are never used to select features, preventing circularity.

\textbf{Aggregation into alignment metrics}
To summarize predictive alignment, we follow the exact aggregation used for SRDM-RSA pipeline and therefore reuse the definitions of $S(m,b)$ and $S(m,b_r)$ introduced in Section~\ref{alignment_pipeline}. The only substitution is that SRDM correlations $\rho(L_j,C_i)$ are replaced by the predictive scores $P(L_j,C_i)$. Concretely, for each layer $L_j$ we rank channels by $P(L_j,C_i)$, take the top 100, and compute $S(m,b)$ and $S(m,b_r)$ by the same layer-averaging rules as before. For the \textbf{sentence} corpus, we compute these summaries separately for sentence, phrase and sentence \& phrase neurons and then report their arithmetic mean to yield a single value per model and hemisphere. For the \textbf{phrase} corpus, the selection procedure yields only phrase-selective neurons; consequently no averaging across syntactic neuron types is required. This results in one model--brain score $S(m,b)$ and one score per ROI $S(m,b_r)$ for each model in each hemisphere, directly comparable to the SRDM results while relying on predictive encoding rather than RSA.

\begin{table}[ht]
\centering
\caption{Predictive alignment results under the \textbf{sentence} corpus. The first row reports the overall model--brain similarity \(S(m,b)\); subsequent rows report the model--region similarity \(S(m,b_r)\). Bold in the \(S(m,b)\) row marks the single highest value across models/hemispheres; within each model, bold highlights the three highest \(S(m,b_r)\) entries across all ROIs and hemispheres.}
\label{predictive_sentence}
\begin{tabularx}{\textwidth}{@{}c*{12}{>{\centering\arraybackslash}X}@{}}
\toprule
 & \multicolumn{2}{c}{GPT-2} & \multicolumn{2}{c}{Gemma} & \multicolumn{2}{c}{Gemma 2} & \multicolumn{2}{c}{Llama 2} & \multicolumn{2}{c}{Llama 3.1} & \multicolumn{2}{c}{GLM-4} \\ 
\cmidrule(lr){2-3}\cmidrule(lr){4-5}\cmidrule(lr){6-7}\cmidrule(lr){8-9}\cmidrule(lr){10-11}\cmidrule(lr){12-13}
 & L & R & L & R & L & R & L & R & L & R & L & R \\
\midrule
$S(m,b)$ & \textbf{0.487} & 0.371 & 0.467 & 0.364 & 0.469 & 0.371 & 0.472 & 0.364 & 0.463 & 0.365 & 0.471 & 0.367 \\
\midrule
A1 & \textbf{0.515} & 0.375 & \textbf{0.473} & 0.368 & \textbf{0.474} & 0.360 & \textbf{0.470} & 0.388 & \textbf{0.470} & 0.360 & \textbf{0.485} & 0.376 \\ 
STG & \textbf{0.499} & 0.384 & \textbf{0.469} & 0.357 & \textbf{0.472} & 0.369 & \textbf{0.480} & 0.369 & 0.464 & 0.373 & \textbf{0.476} & 0.374 \\ 
MTG & 0.487 & 0.370 & 0.468 & 0.366 & \textbf{0.470} & 0.375 & 0.475 & 0.369 & \textbf{0.465} & 0.366 & 0.469 & 0.371 \\ 
ITG & 0.479 & 0.370 & 0.463 & 0.361 & 0.466 & 0.367 & 0.466 & 0.360 & 0.457 & 0.362 & 0.466 & 0.364 \\ 
Insula & 0.486 & 0.373 & 0.459 & 0.368 & 0.468 & 0.377 & \textbf{0.476} & 0.376 & 0.463 & 0.363 & \textbf{0.473} & 0.370 \\ 
TPJ & 0.474 & 0.345 & 0.468 & 0.349 & 0.466 & 0.361 & 0.463 & 0.347 & 0.462 & 0.347 & 0.467 & 0.350 \\ 
Temporal Pole & 0.484 & 0.384 & 0.467 & 0.362 & 0.455 & 0.373 & 0.475 & 0.365 & 0.459 & 0.362 & 0.469 & 0.351 \\ 
Sensorimotor & \textbf{0.491} & 0.373 & \textbf{0.470} & 0.363 & 0.471 & 0.375 & 0.470 & 0.362 & \textbf{0.466} & 0.366 & 0.472 & 0.370 \\ 
IFG & 0.480 & 0.370 & 0.465 & 0.365 & 0.465 & 0.366 & 0.466 & 0.360 & 0.462 & 0.359 & 0.467 & 0.364 \\ 
MFG & 0.475 & 0.367 & 0.457 & 0.374 & 0.462 & 0.364 & 0.466 & 0.355 & 0.453 & 0.373 & 0.462 & 0.376 \\ 
Hippocampus & 0.481 & 0.358 & 0.464 & 0.365 & 0.458 & 0.366 & 0.462 & 0.353 & 0.458 & 0.360 & 0.469 & 0.362 \\ 
Amygdala & 0.472 & 0.334 & 0.463 & 0.345 & 0.465 & 0.362 & 0.452 & 0.353 & 0.463 & 0.358 & 0.471 & 0.349 \\ 
\bottomrule
\end{tabularx}
\vskip -0.1in
\end{table}

\begin{table}[ht]
\centering
\caption{Predictive alignment results under the \textbf{phrase} corpus. The first row reports the overall model--brain similarity $S(m,b)$; each subsequent brain ROI row reports the model--region similarity $S(m,b_r)$. Bold in the \(S(m,b)\) row marks the single highest value across models/hemispheres; within each model, bold highlights the three highest \(S(m,b_r)\) entries across all ROIs and hemispheres.}
\label{predictive_phrase}
\begin{tabularx}{\textwidth}{@{}c*{13}{>{\centering\arraybackslash}X}@{}}
\toprule
 & \multicolumn{2}{c}{GPT-2} & \multicolumn{2}{c}{Gemma} & \multicolumn{2}{c}{Gemma 2} & \multicolumn{2}{c}{Llama 2} & \multicolumn{2}{c}{Llama 3.1} & \multicolumn{2}{c}{GLM-4} \\
\cmidrule(lr){2-13}
 & L & R & L & R & L & R & L & R & L & R & L & R \\
\midrule
$S(m,b)$ & 0.454 & 0.356 & \textbf{0.465} & 0.362 & \textbf{0.465} & 0.366 & 0.446 & 0.350 & 0.451 & 0.354 & 0.449 & 0.349 \\
\midrule
A1 & 0.448 & 0.351 & \textbf{0.478} & 0.343 & \textbf{0.482} & 0.345 & 0.451 & 0.365 & 0.452 & 0.333 & 0.443 & 0.344 \\
STG & \textbf{0.463} & 0.363 & \textbf{0.468} & 0.364 & \textbf{0.470} & 0.363 & 0.448 & 0.356 & \textbf{0.455} & 0.352 & \textbf{0.458} & 0.357 \\
MTG & 0.450 & 0.357 & 0.464 & 0.362 & 0.462 & 0.366 & 0.445 & 0.363 & 0.448 & 0.348 & 0.445 & 0.345 \\
ITG & 0.449 & 0.348 & \textbf{0.465} & 0.359 & 0.460 & 0.363 & 0.445 & 0.346 & 0.448 & 0.352 & 0.448 & 0.345 \\
Insula & 0.453 & 0.360 & 0.461 & 0.360 & 0.462 & 0.365 & 0.446 & 0.348 & 0.451 & 0.354 & \textbf{0.452} & 0.344 \\
TPJ & 0.454 & 0.350 & 0.462 & 0.361 & 0.460 & 0.361 & 0.444 & 0.343 & \textbf{0.450} & 0.350 & 0.448 & 0.349 \\
Temporal Pole & \textbf{0.461} & 0.371 & 0.464 & 0.363 & \textbf{0.464} & 0.365 & \textbf{0.450} & 0.347 & 0.448 & 0.354 & \textbf{0.452} & 0.344 \\
Sensorimotor & \textbf{0.461} & 0.371 & 0.464 & 0.363 &\textbf{0.464} & 0.365 & 0.450 & 0.347 & 0.448 & 0.354 & \textbf{0.452} & 0.344 \\
IFG & 0.450 & 0.351 & 0.454 & 0.362 & 0.453 & 0.358 & 0.437 & 0.343 & \textbf{0.454} & 0.353 & 0.448 & 0.347 \\
MFG & 0.451 & 0.328 & 0.466 & 0.333 & 0.460 & 0.361 & \textbf{0.455} & 0.330 & \textbf{0.455} & 0.353 & \textbf{0.457} & 0.351 \\
Hippocampus & \textbf{0.454} & 0.354 & 0.453 & 0.363 & 0.463 & 0.357 & 0.436 & 0.344 &\textbf{0.454} & 0.354 & 0.434 & 0.354 \\
Amygdala & 0.403 & 0.361 & 0.420 & 0.391 & 0.450 & 0.355 & 0.385 & 0.349 & 0.402 & 0.386 & 0.389 & 0.381 \\
\bottomrule
\end{tabularx}
\vskip -0.1in
\end{table}

Across both corpora (Tables \ref{predictive_sentence} and \ref{predictive_phrase}), a one-way ANOVA on layer-averaged top-100 predictive scores revealed reliable between-model differences in brain-level alignment \(S(m,b)\) (\textbf{sentence}: \(F=43.37,\,p<0.001,\,\eta^2=0.008\); \textbf{phrase}: \(F=13.80,\,p<0.001,\,\eta^2=0.008\)), with consistently higher alignment in the left hemisphere. At the region level, FDR-controlled one-way ANOVAs showed significant between-model effects within left-dominant language cortices (e.g., STG, MTG, ITG) for both corpora. These findings align with the SRDM–RSA analyses, which likewise demonstrate significant between-model variation at both whole-brain and region-specific levels.

Mirroring the SRDM–RSA trend, the predictive results exhibit an almost identical trend for model upgrade comparison. In the \textbf{sentence} corpus, the left-hemisphere model–brain scores reproduce the RSA ordering—Gemma 2 exceeds Gemma (\(S(m,b)=0.469\) vs.\ \(0.467\), L), while Llama 3.1 falls below Llama 2 (\(0.463\) vs.\ \(0.472\), L). In the \textbf{phrase} corpus, the gaps compress: Gemma 2 is essentially tied with Gemma (both \(S(m,b)=0.465\), L), and Llama 3.1 is slightly higher than Llama 2 (\(0.451\) vs.\ \(0.446\), L). This compression likely arises because phrase analyses use only phrase neurons and provide fewer, noisier frequency samples, which together reduce model separability. Taken together, the predictive encoding and SRDM–RSA analyses converge on a consistent, method-robust conclusion: inter-model differences are statistically reliable at both the brain and region levels, and, at the brain level, model-upgrade trends are essentially identical across methods, with only minor fluctuations.

\section{Bilingual sentence- and phrase-level representations in LLMs}
\label{multilingual}

Previous studies have explored how LLMs handle different languages, concluding that while most neurons are shared across languages, a smaller subset of neurons is dedicated to processing specific languages \citep{tangLanguageSpecificNeuronsKey2024}. But does this hold true for syntactic structure perception? This appendix provides insights into this question. It is important to note that the GPT-2 model used in this study was pre-trained on a Chinese corpus using the Universal Encoder Representations (UER) framework \citep{zhaoUEROpenSourceToolkit2019}, equipping it with the capability to process both Chinese and English text effectively. This stands in contrast to the original GPT-2 model \citep{radford2019language}, which lacks the ability to handle Chinese text.

\begin{figure}[ht]
    \centering
    % First row
    \begin{subfigure}{0.47\linewidth}
        \includegraphics[width=\linewidth]{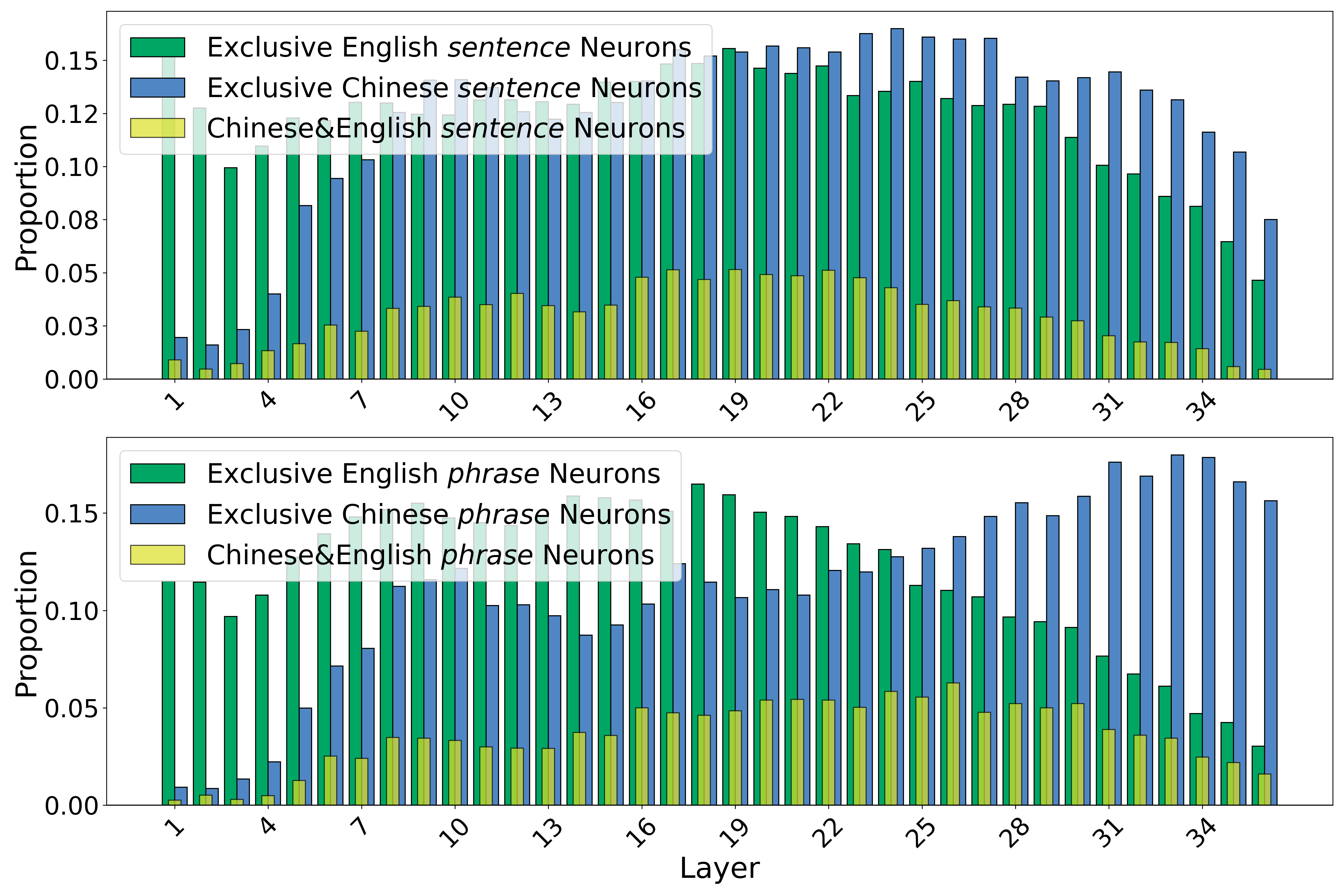}
        \caption{GPT-2}
        \label{fig:gpt-2}
    \end{subfigure}
    \hfill
    \begin{subfigure}{0.47\linewidth}
        \includegraphics[width=\linewidth]{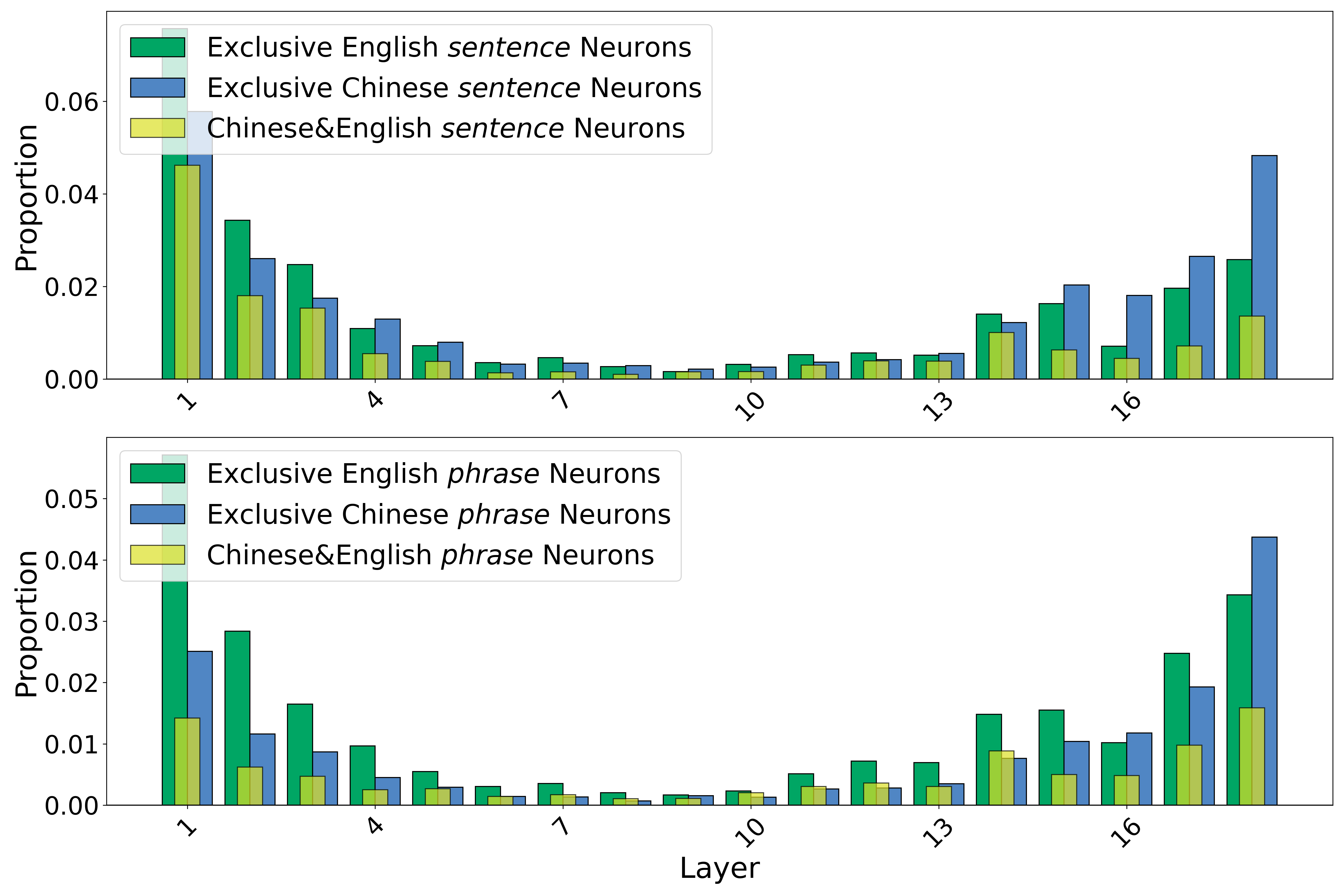}
        \caption{Gemma}
        \label{fig:gemma}
    \end{subfigure}

    \vspace{1em} % Add vertical spacing between rows
    % Second row
    \begin{subfigure}{0.47\linewidth}
        \includegraphics[width=\linewidth]{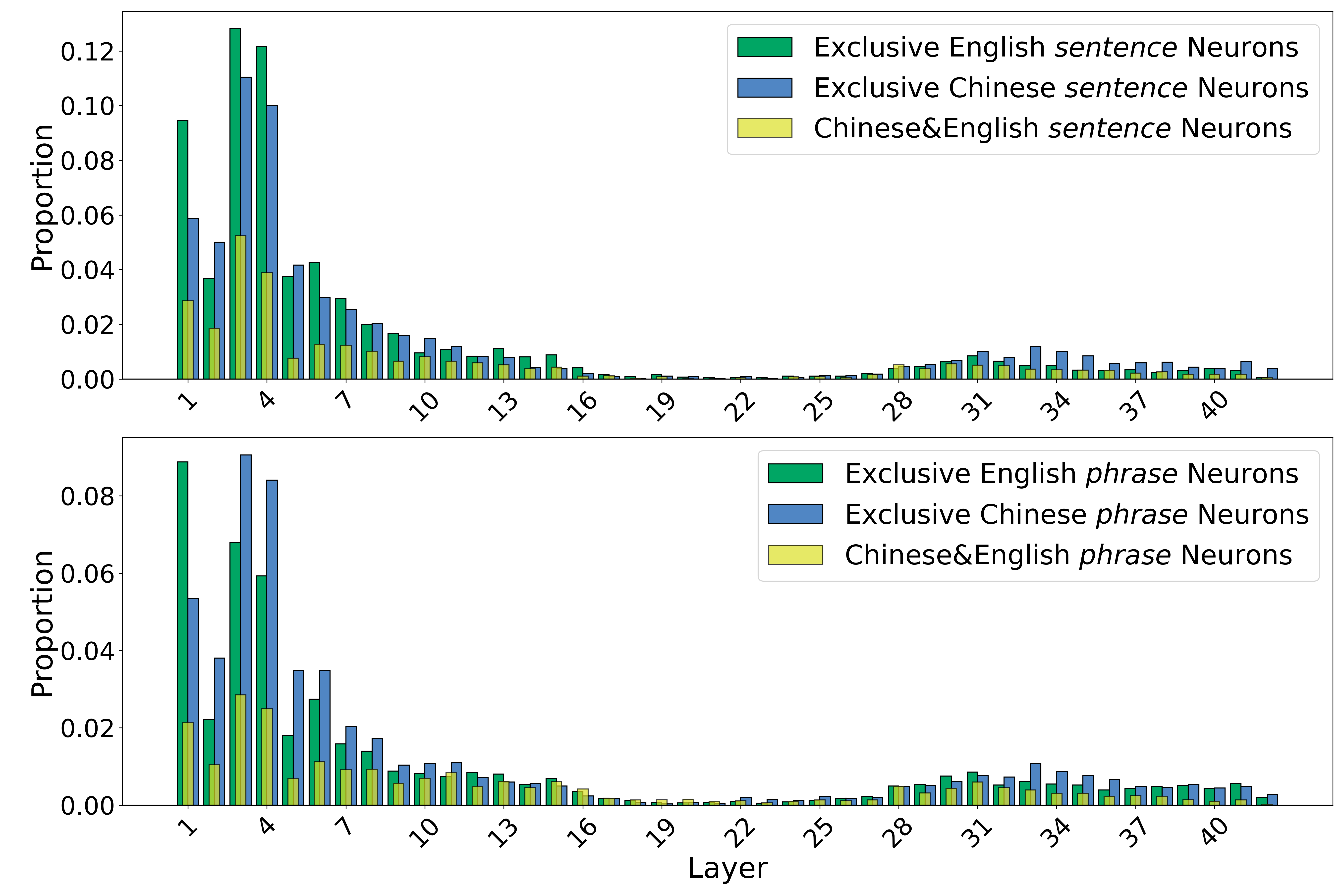}
        \caption{Gemma 2}
        \label{fig:gemma2}
    \end{subfigure}
    \hfill
    \begin{subfigure}{0.47\linewidth}
        \includegraphics[width=\linewidth]{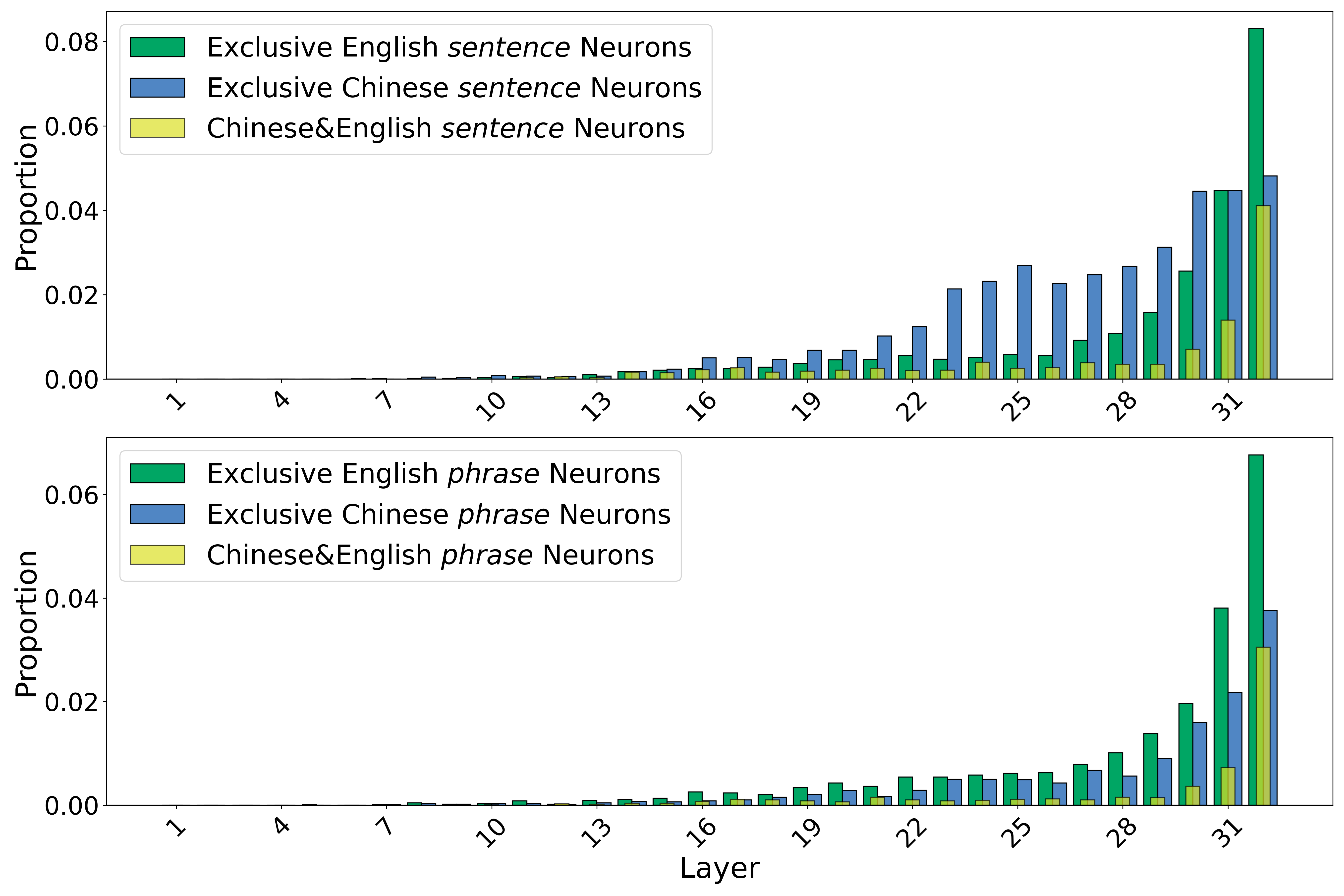}
        \caption{Llama 2}
        \label{fig:llama2}
    \end{subfigure}

\vspace{1em} % Add vertical spacing between rows
    % Second row
    \begin{subfigure}{0.47\linewidth}
        \includegraphics[width=\linewidth]{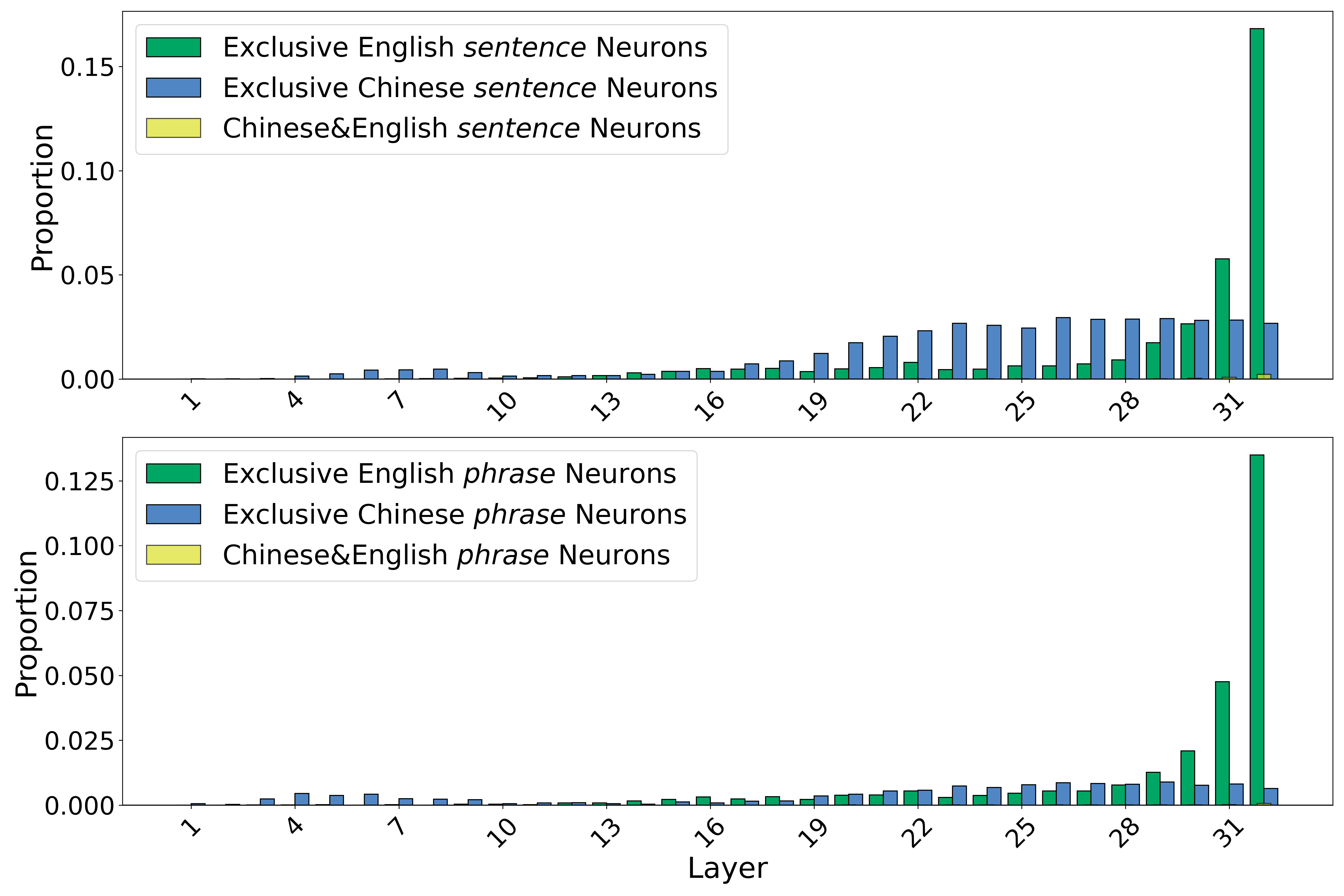}
        \caption{Llama 3.1}
        \label{fig:llama3}
    \end{subfigure}
    \hfill
    \begin{subfigure}{0.47\linewidth}
        \includegraphics[width=\linewidth]{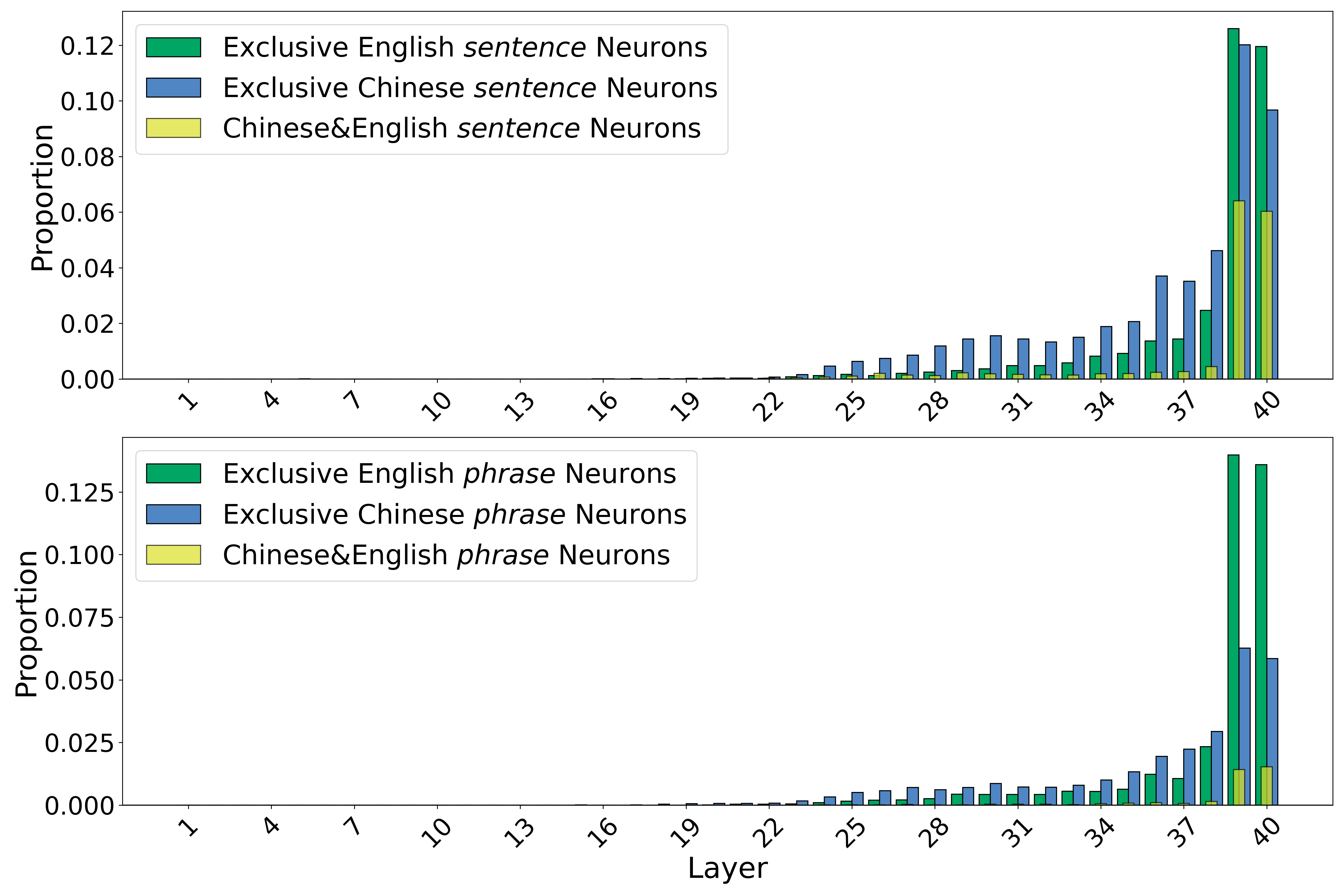}
        \caption{GLM-4}
        \label{fig:glm}
    \end{subfigure}

    \caption{Cross-language neural representations extracted from five multilingual models (Gemma, Gemma 2, Llama 2, Llama 3.1, and GLM-4) depicting syntactic processing capabilities.}
    \label{cross-language-neurons}
\end{figure}

The results in Figure \ref{cross-language-neurons} suggest that language-specific syntactic neurons (i.e., exclusive sentence/phrase neurons) tend to cluster toward the final layers of Llama 2, Llama 3.1, and GLM-4, with the proportion of bilingual neurons (Chinese \& English) increasing progressively in deeper layers. In contrast, Gemma and Gemma 2 display different patterns. In Gemma, both language-specific and bilingual neurons are found not only in the deeper layers but also in the initial layers, whereas this is only observed in the early layers of Gemma 2. GPT-2, on the other hand, exhibits a more balanced distribution, with both language-specific and bilingual neurons present in almost every layer in roughly equal proportions. Furthermore, English-specific neurons are more prominent in the early layers, while Chinese-specific neurons are more concentrated in the later layers. Interestingly, Llama 3.1 shows a notably lower count of Chinese-specific neurons compared to English-specific neurons in the final layer, and fewer Chinese \& English neurons than the other five LLMs. Although Llama 3.1 was pre-trained on 176 languages \citep{dubey2024llama}, it appears to have less specialization in Chinese, which may explain the reduced presence of Chinese-specific neurons and, consequently, fewer bilingual neurons.

\section{Preliminary bilingual HFTP test on native Chinese speakers}
\label{multilingual_human}

\begin{figure*}[ht]
\begin{center}
\centerline{\includegraphics[width=\linewidth]{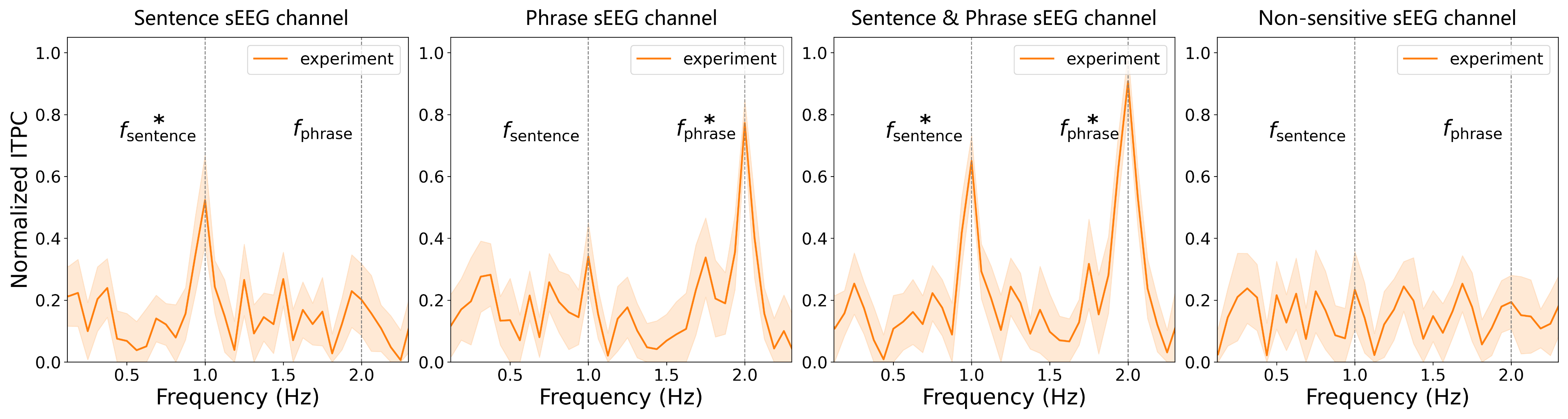}}
 \caption{Hierarchical frequency patterns of sEEG channels from one participant, using four-word English sequence, selectively represent sentence features, phrase features, shared features of both  and non-sensitive feature (from left to the right). Shaded bands show \(\pm 1\;\mathrm{s.e.m.}\) computed within each channel by bootstrapping across sliding-window ITPC estimates.}
    \label{itpc_eng}
\end{center}
\vskip -0.2in
\end{figure*}

As noted above, our HFTP analysis was originally conducted on native Chinese speakers listening to a Chinese corpus. To probe cross-linguistic generalization in human sEEG, we constructed bilingual materials by taking the Chinese corpus from \citep{sheng2019cortical} and manually translating it into four-word English sequences. We report preliminary results from one native Chinese speaker who listened to both corpora. Hierarchical frequency patterns for the Chinese corpus closely match those in Figure \ref{itpc}, and the corresponding English results are shown in Figure \ref{itpc_eng}.

Notably, the English condition exhibits robust sentence- (\(1\;\mathrm{Hz}\)) and phrase-rate (\(2\;\mathrm{Hz}\)) peaks, mirroring the Chinese condition and indicating alignment to hierarchical syntactic structure beyond the native language of the participant. Building on this convergence, we hypothesize that English speakers who also understand Chinese will likewise exhibit sentence- and phrase-rate peaks under the same experimental setting. The concordance of spectra motivates a cross-linguistic interpretation in which the \(1\) and \(2\;\mathrm{Hz}\) signature reflects domain-general constraints on chunking and predictive parsing rather than language-specific lexical statistics. Hence cross-linguistic factors such as function-word density may redistribute spectral power but should not abolish the peaks. This view also aligns with Ding et al. \citep{dingCorticalTrackingHierarchical2016}, who showed that the syntactic structure hierarchy is a language-general signature of syntactic grouping, independent of word-order typology or writing system.

\section{Contribution Ratios of LLMs}
\label{contri_llm}

To further investigate the role of specific brain ROIs in syntactic processing, we introduced the contribution ratio ($CR_{r}$). The contribution ratio highlights which brain ROIs contribute most significantly to the syntactic alignment between LLMs and the human brain. Fixing a model layer, this metric quantifies the influence of each brain ROIs by calculating the proportion of channels from a given region within the top 100 most relevant channels, normalized by the overall representation of the ROIs (results can be found in Appendix \ref{contri_llm}). The contribution ratio is defined as:

\begin{equation} CR_{r}(L_j) = \frac{N_r^{\text{top}}(L_j) / N^{\text{top}}}{N_r^{\text{total}}/ N^{\text{total}}} ,\end{equation}

where $N_r^{\text{top}}(L_j)$ is the number of channels in region $r$ within the top 100 channels in terms of the LLM layer $L_j$, $N^{\text{top}}$ is the total number of top channels, which is specified as $100$ in this case, $N_r^{\text{total}}$ is the total number of channels in region $r$,  and $N^{\text{total}}$ is the total number of brain channels.

\begin{figure}[h]
    \centering
    \includegraphics[width=1\linewidth]{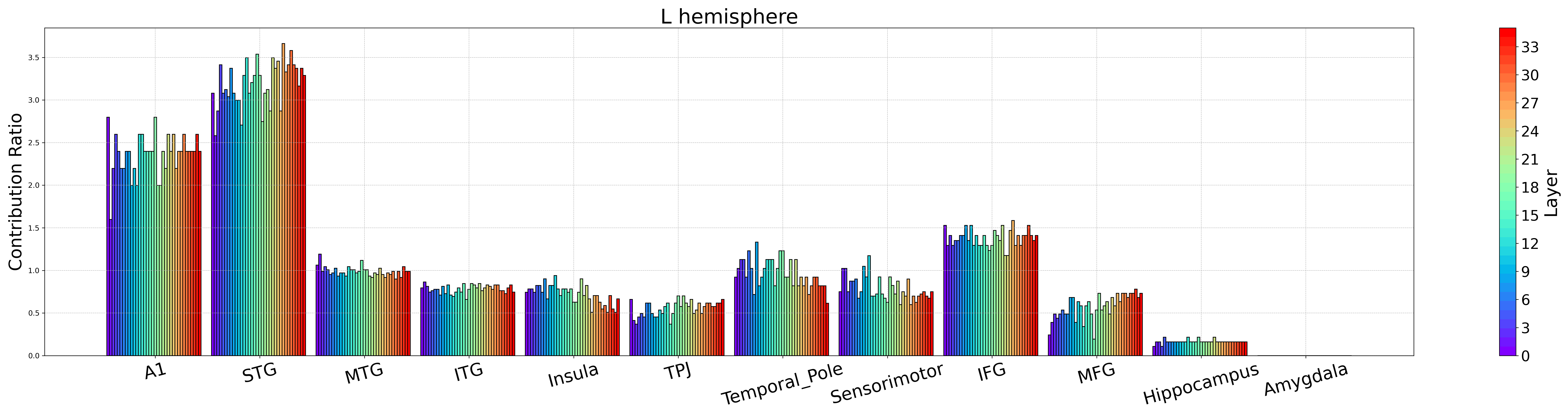}
    \includegraphics[width=1\linewidth]{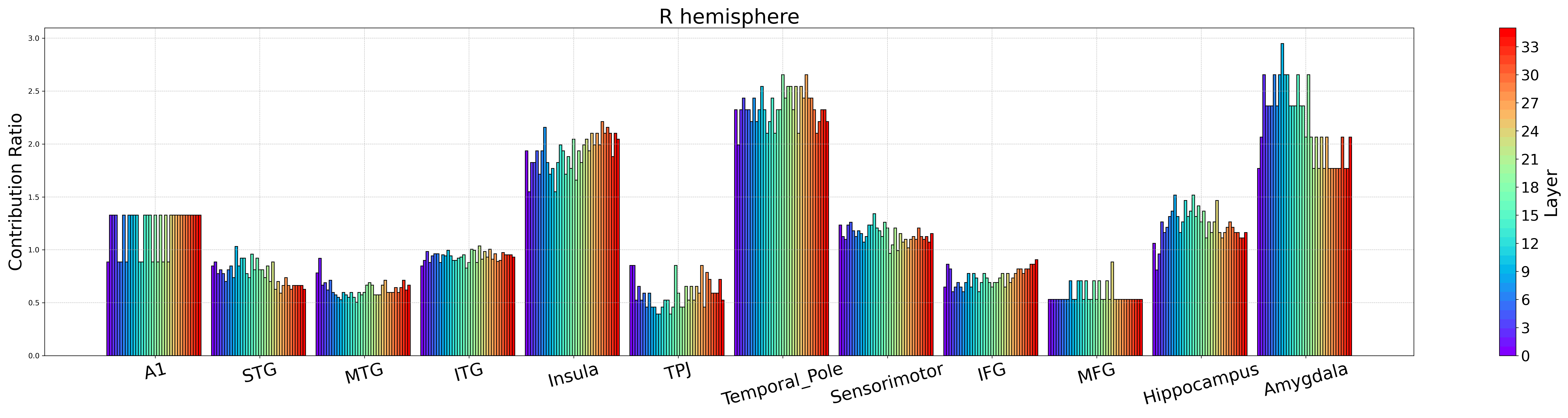}
    \caption{Contribution ratios for GPT-2: Left hemisphere (top) and Right hemisphere (bottom).}
    \label{gpt2-contribution}
\end{figure}

\begin{figure}[h]
    \centering
    \includegraphics[width=1\linewidth]{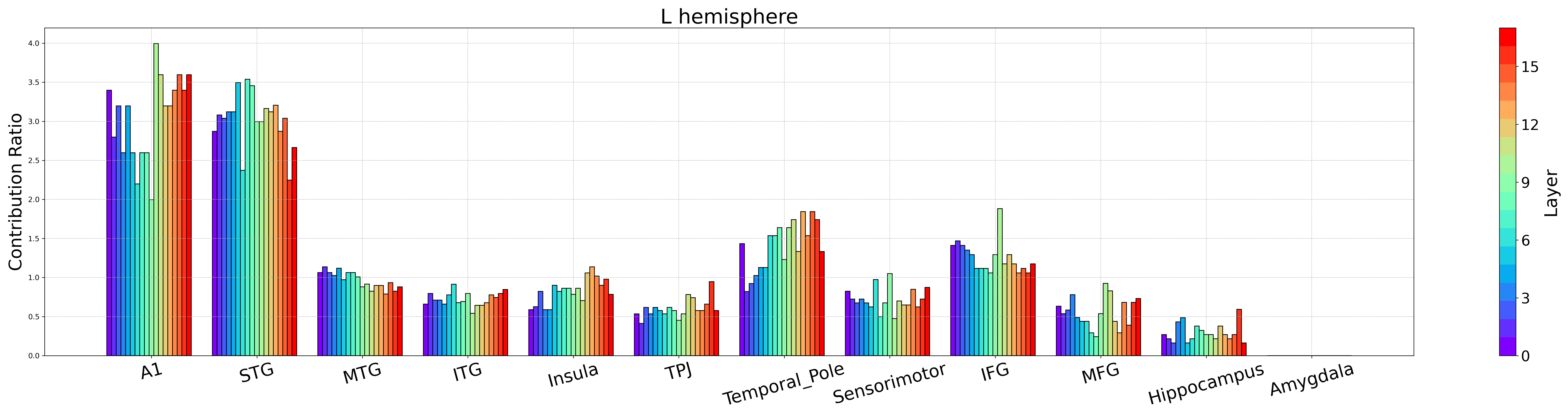}
    \includegraphics[width=1\linewidth]{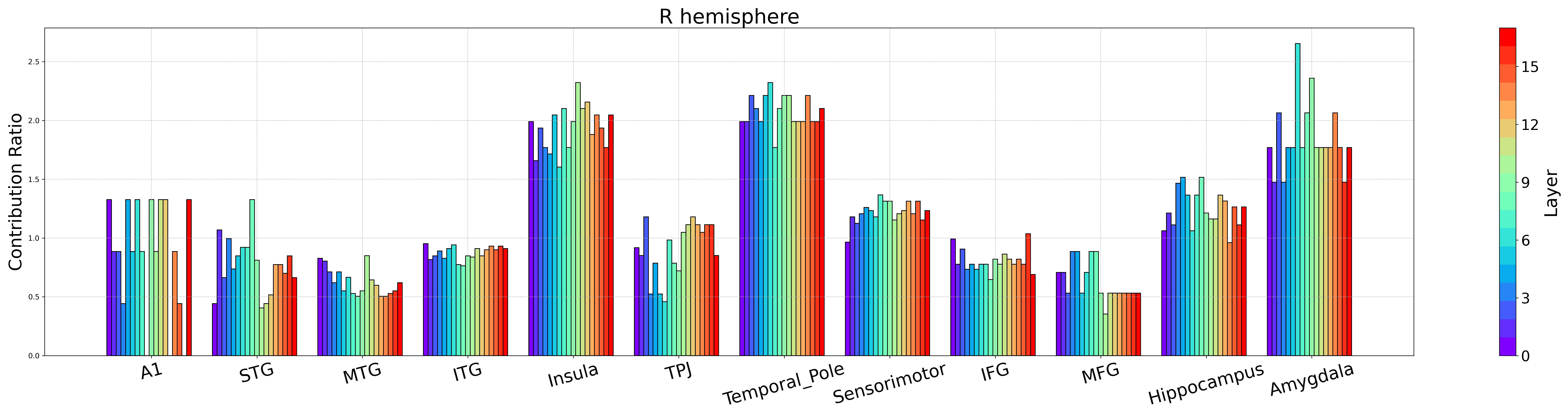}
    \caption{Contribution ratios for Gemma: Left hemisphere (top) and Right hemisphere (bottom).}
    \label{gemma-contribution}
\end{figure}

\begin{figure}[h]
    \centering
    \includegraphics[width=1\linewidth]{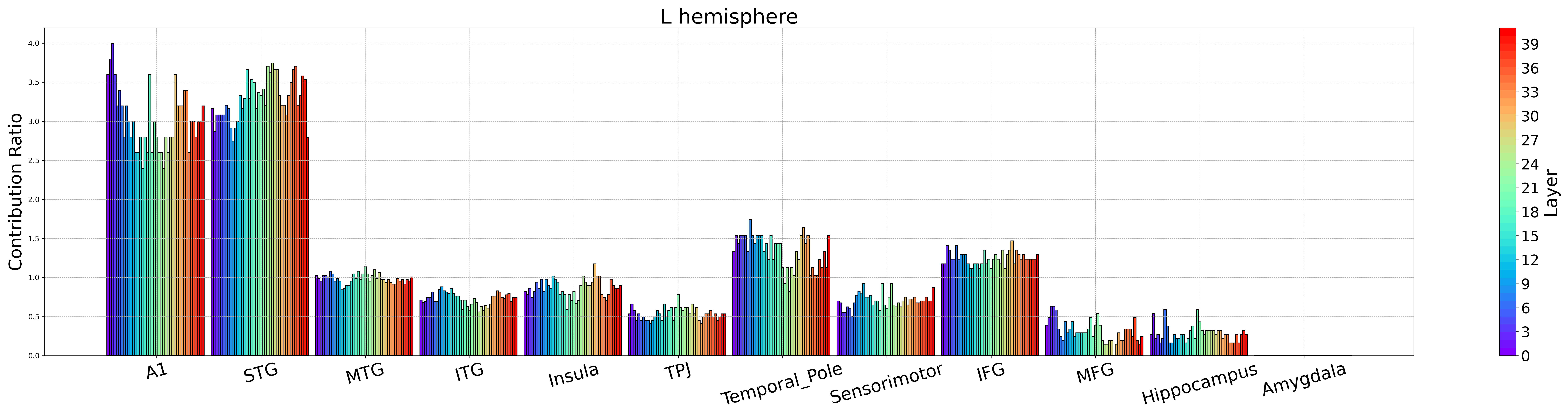}
    \includegraphics[width=1\linewidth]{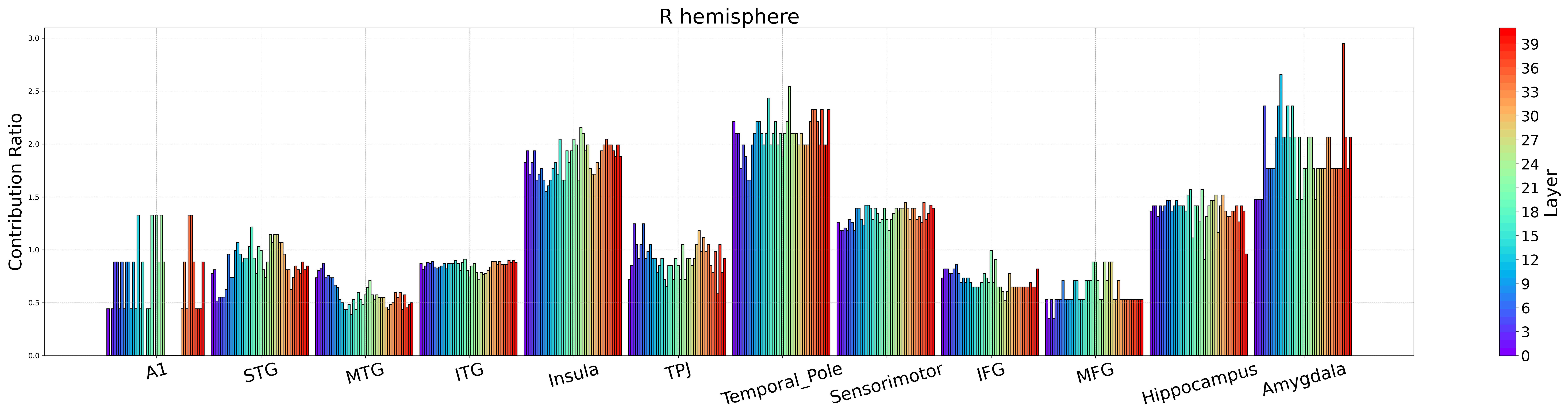}
    \caption{Contribution ratios for Gemma 2: Left hemisphere (top) and Right hemisphere (bottom).}
    \label{gemma2-contribution}
\end{figure}

\begin{figure}[h]
    \centering
    \includegraphics[width=1\linewidth]{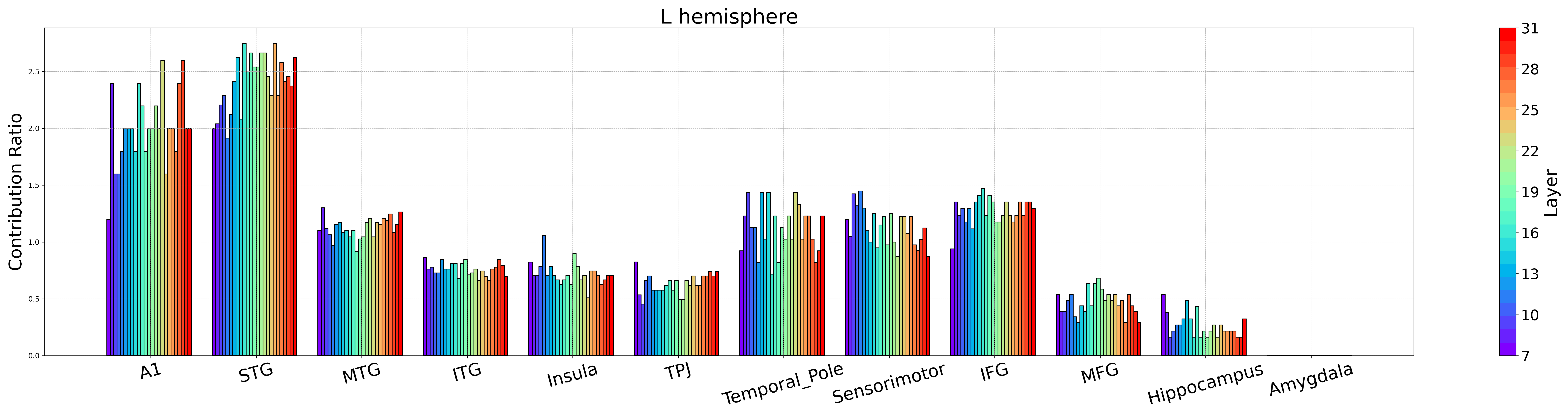}
    
    \includegraphics[width=1\linewidth]{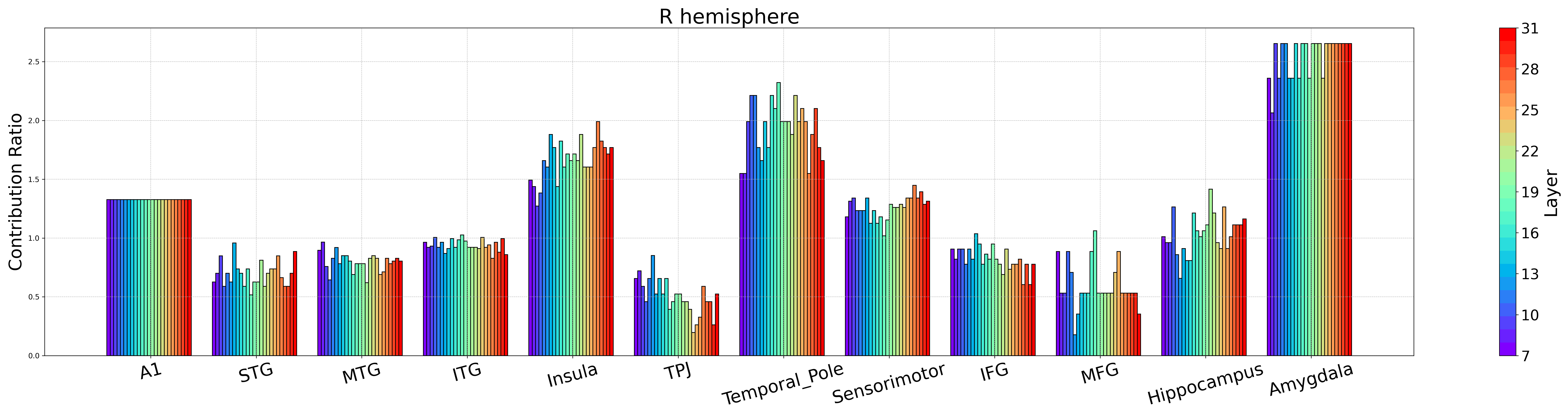}
    
    \caption{Contribution ratios for Llama 2: Left hemisphere (top) and Right hemisphere (bottom).}
    \label{contribution}
\end{figure}

\begin{figure}[h]
    \centering
    \includegraphics[width=1\linewidth]{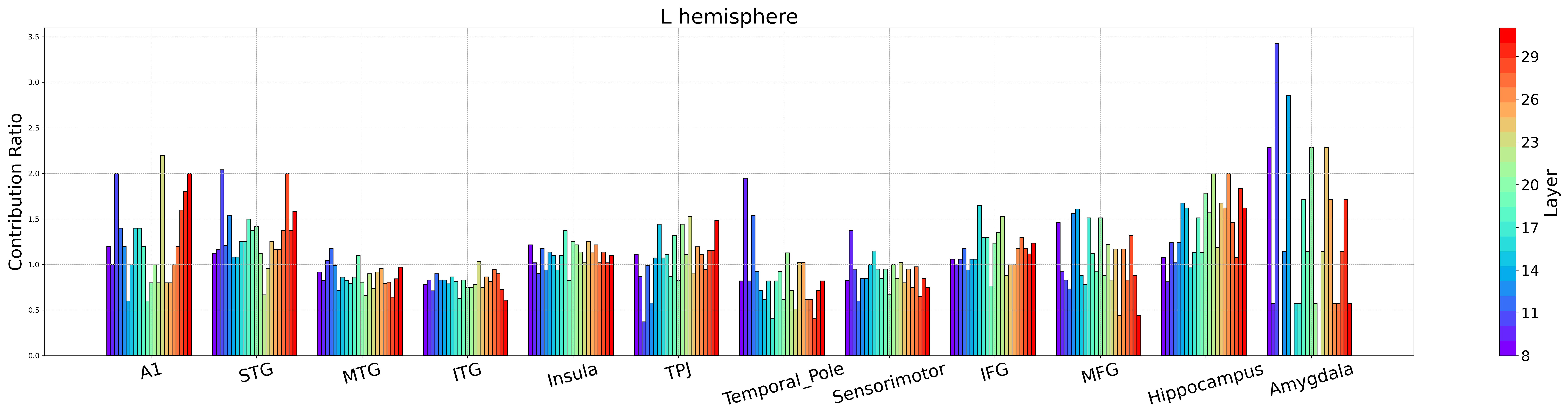}
    \includegraphics[width=1\linewidth]{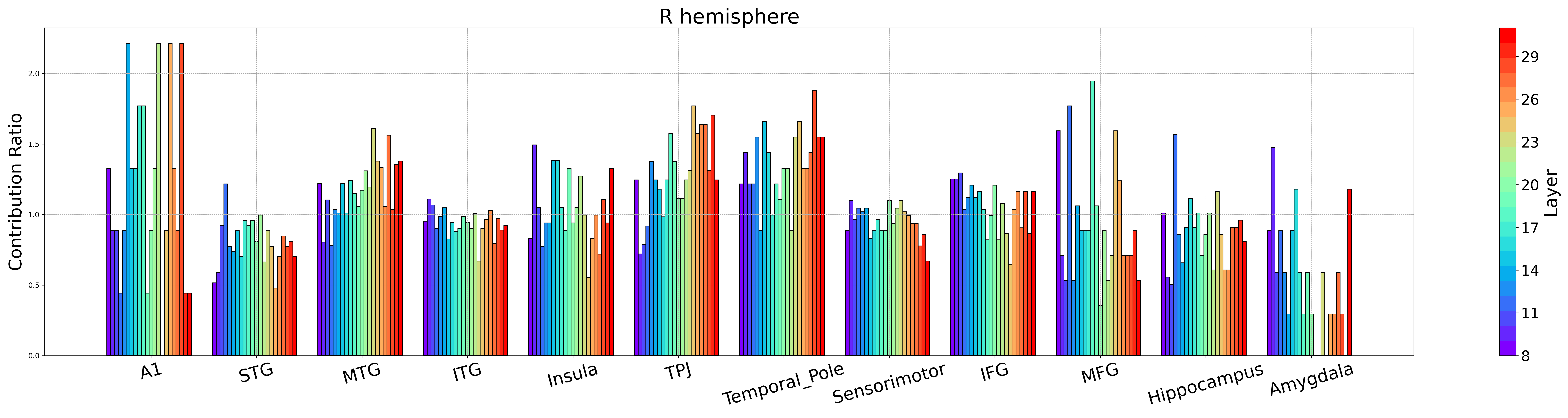}
    \caption{Contribution ratios for Llama 3.1: Left hemisphere (top) and Right hemisphere (bottom).}
    \label{llama-3.1-contribution}
\end{figure}

\begin{figure}[h]
    \centering
    \includegraphics[width=1\linewidth]{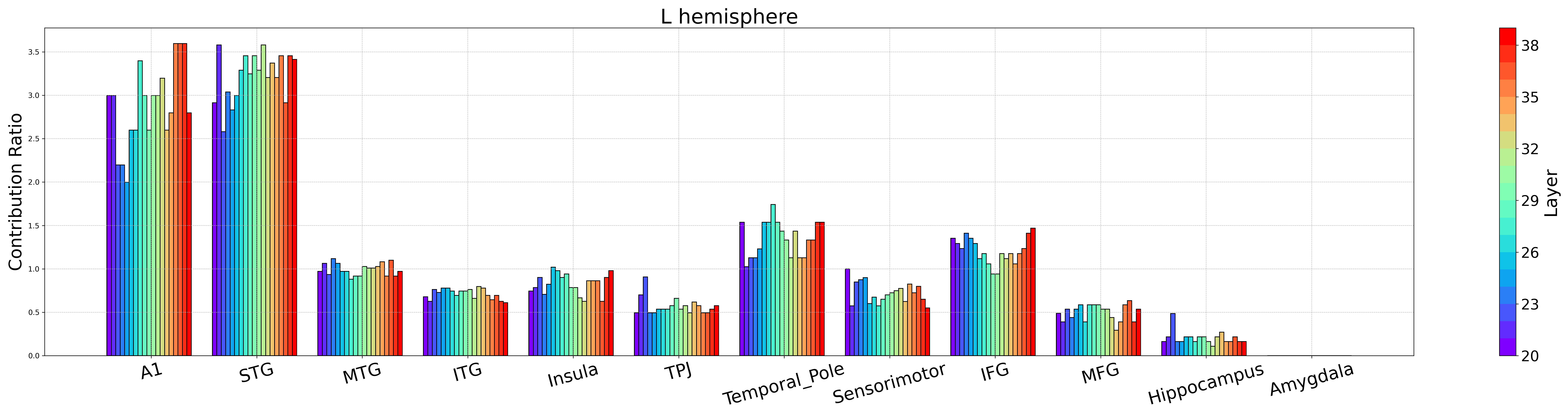}
    \includegraphics[width=1\linewidth]{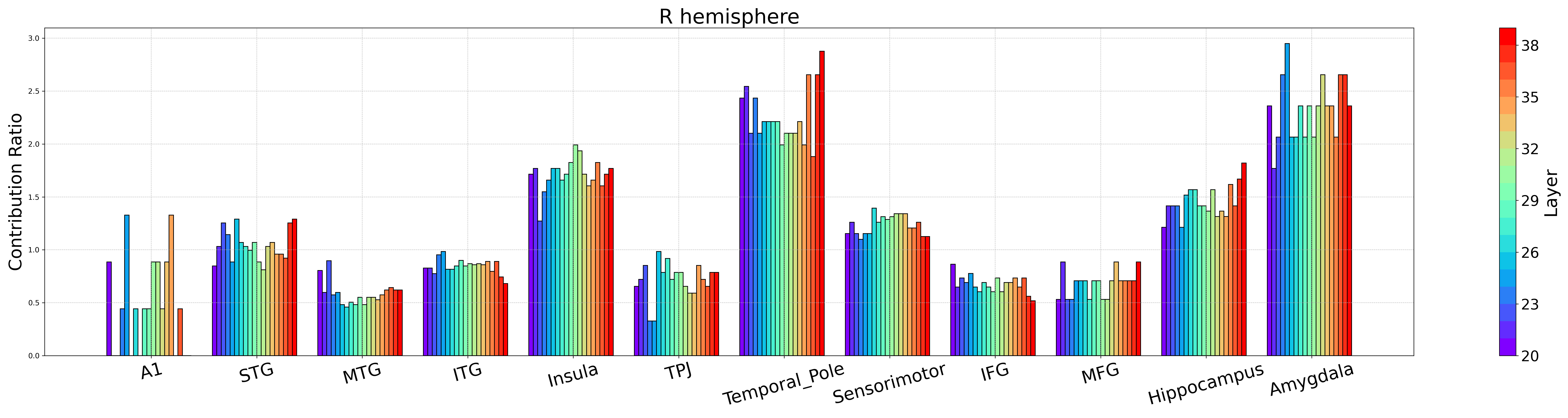}
    \caption{Contribution ratios for GLM-4: Left hemisphere (top) and Right hemisphere (bottom).}
    \label{glm-4-contribution}
\end{figure}

In this appendix we present the contribution ratio results for six LLMs used in this study: GPT-2, Gemma, Gemma 2, Llama 2, Llama 3.1, and GLM-4. Specifically, the contribution ratio for each model was calculated based on the number of top 100 significant channels within each brain ROIs, as described in Appendix \ref{alignment_pipeline}. Below, we present the results for both the left (L) and right (R) hemispheres of each model (See Figures \ref{gpt2-contribution}, \ref{gemma-contribution}, \ref{gemma2-contribution}, \ref{contribution}, \ref{llama-3.1-contribution} and 
\ref{glm-4-contribution}). These figures offer further insights into how different LLMs align with human brain ROIs in terms of syntactic processing.

From these figures, we observe that across all LLMs, regions such as A1 and STG in the left hemisphere, and the Insula, Temporal Pole, and Amygdala in the right hemisphere contribute more significantly to the alignment with human brain syntactic processing. These regions are known to be involved in language-specific processes in the human brain, particularly in the left hemisphere, where the STG and A1 are crucial for auditory and syntactic processing. The alignment results suggest that these models may be capturing aspects of hierarchical syntactic structures in ways that are functionally similar to human neural mechanisms. The Insula, Temporal Pole, and Amygdala, though not traditionally highlighted as primary language regions, may also play supporting roles in language comprehension, possibly through emotion and memory-related pathways. This suggests that LLMs might engage both language-specific and auxiliary brain ROIs to process syntax, mirroring the integrated and distributed nature of human brain networks involved in language processing.

\section{HFTP on naturalistic corpus}
\label{natural}

\begin{figure}[ht]
    \centering
    \includegraphics[width=\linewidth]{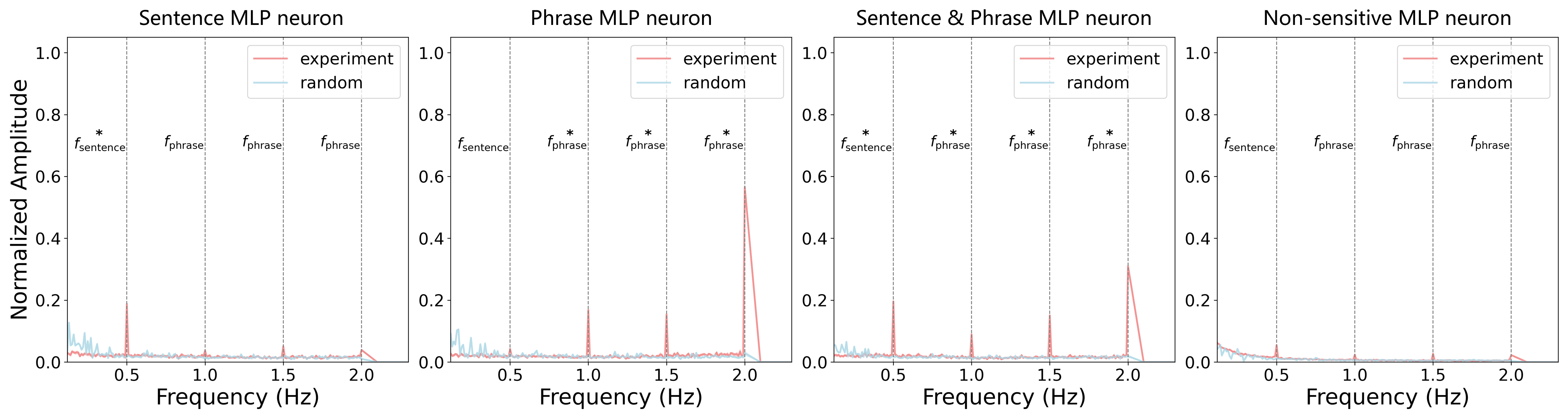}
    \caption{Hierarchical frequency patterns of MLP neurons, using a naturalistic Chinese 8-syllable corpus, selectively represent sentence features, phrase features, shared features of both, and non-sensitive features (from left to right).}
    \label{natural8c}
\end{figure}

\begin{figure}[ht]
    \centering
    \includegraphics[width=\linewidth]{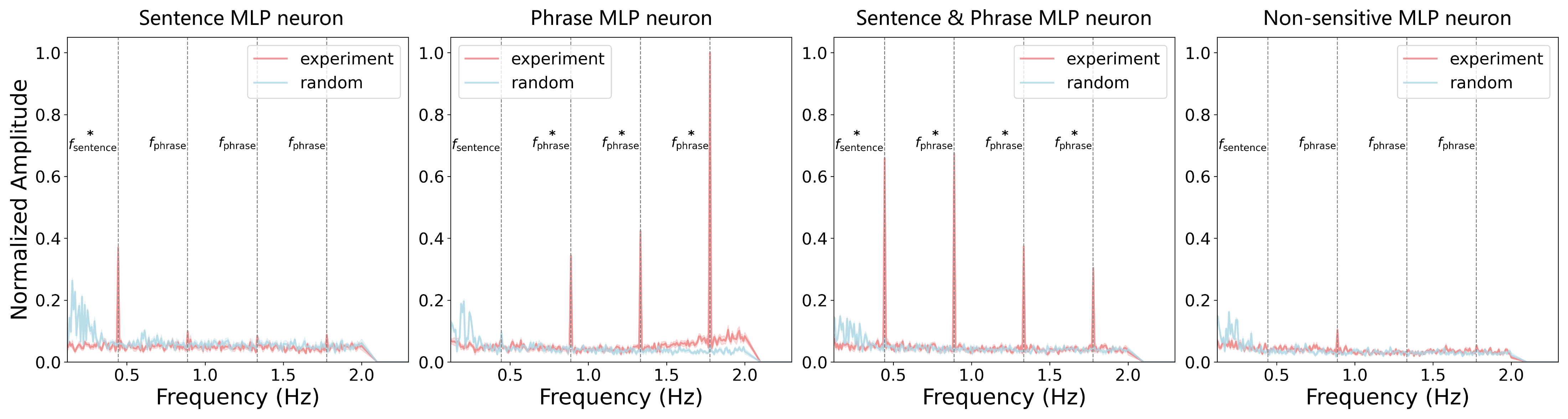}
    \caption{Hierarchical frequency patterns of MLP neurons, using a naturalistic Chinese 9-syllable corpus, selectively represent sentence features, phrase features, shared features of both, and non-sensitive features (from left to right).}
    \label{natural9c}
\end{figure}

\begin{figure}[ht]
    \centering
    \includegraphics[width=\linewidth]{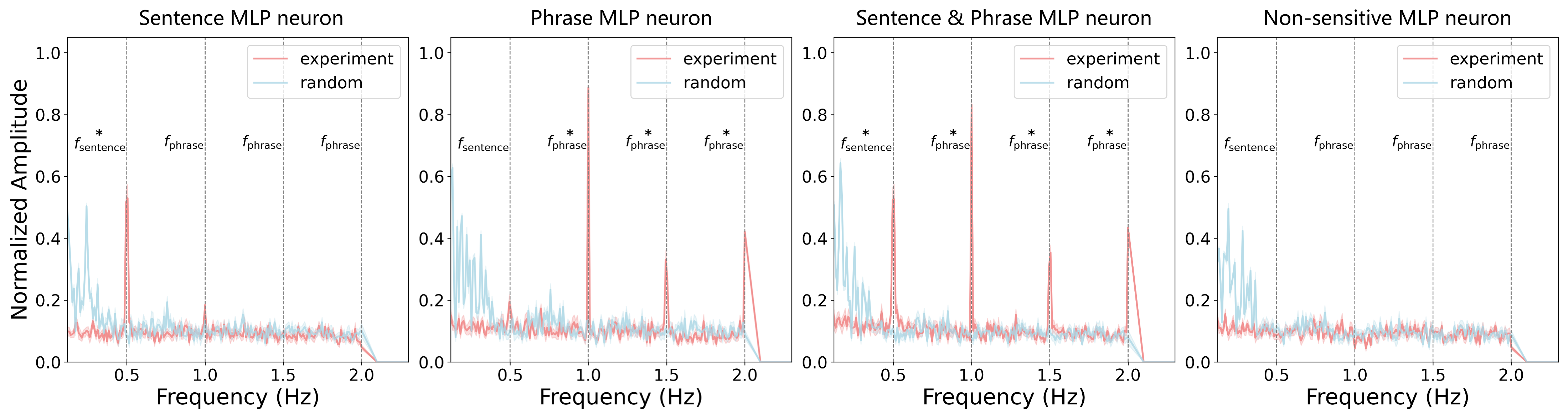}
    \caption{Hierarchical frequency patterns of MLP neurons, using a naturalistic English 8-word corpus, selectively represent sentence features, phrase features, shared features of both, and non-sensitive features (from left to right).}
    \label{natural8e}
\end{figure}

\begin{figure}[ht]
    \centering
    \includegraphics[width=\linewidth]{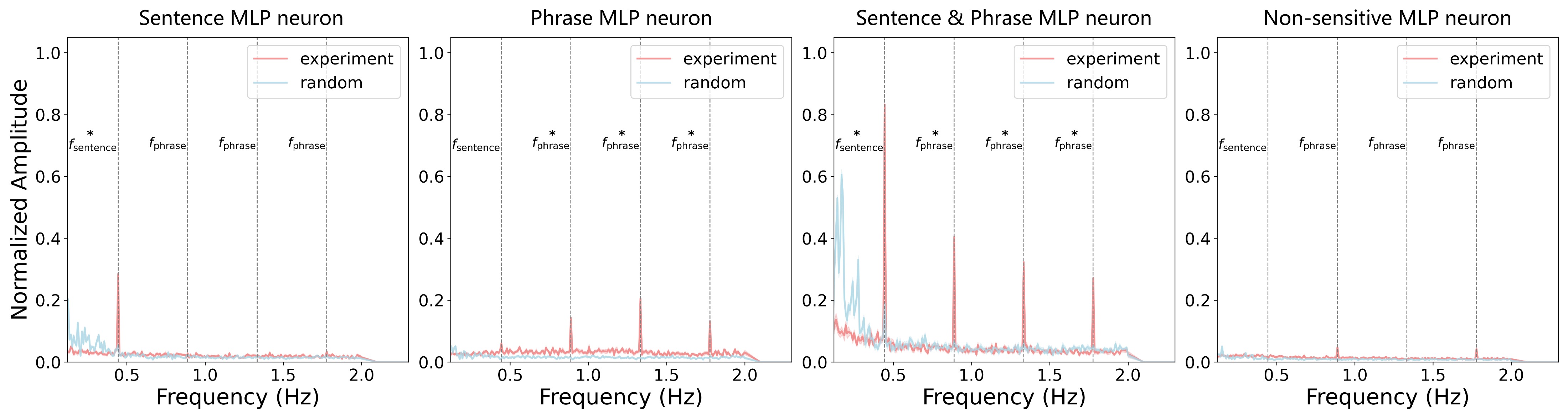}
    \caption{Hierarchical frequency patterns of MLP neurons, using a naturalistic English 9-word corpus, selectively represent sentence features, phrase features, shared features of both, and non-sensitive features (from left to right).}
    \label{natural9e}
\end{figure}

Apart from validating HFTP with highly–controlled four–syllable/word sequences, we show that this probe generalises smoothly to naturalistic text drawn from everyday dialogue, news reports, literary excerpts, poetry, etc.  In these eight- and nine-syllable Chinese corpora and in matched English 8- and 9-word corpora (Figures \ref{natural8c}–\ref{natural9c}, \ref{natural8e}–\ref{natural9e}) the method yields four pronounced spectral peaks per condition.  For the eight-syllable set, peaks arise at \(0.5\), \(1.0\), \(1.5\) and \(2.0\;\mathrm{Hz}\), corresponding to the full–sentence envelope, the canonical 4-character phrase rhythm, an intermediate 2–3-beat grouping around \(1.5\;\mathrm{Hz}\), and the ubiquitous 2-character lexical rhythm, with the English 8-word corpus exhibiting the same four-peak pattern at the corresponding sentence and phrase rates (Figure \ref{natural8e}).  In nine-syllable sentences the peaks shift to \(\sim0.44\), \(0.89\), \(1.33\), and \(1.78\;\mathrm{Hz}\): the lowest peak reflects the whole clause, \(1.33\;\mathrm{Hz}\) aligns with abundant 3-character phrases, \(1.78\;\mathrm{Hz}\) captures rapid two-to-three-character alternations, and the \(0.89\;\mathrm{Hz}\) component indexes a prosodic half-sentence “breath group” of four–to–five characters, and the English 9-word corpus shows the analogous four peaks near these frequencies (Figure \ref{natural9e}).

\begin{figure}[ht]
    \centering
    \includegraphics[width=\linewidth]{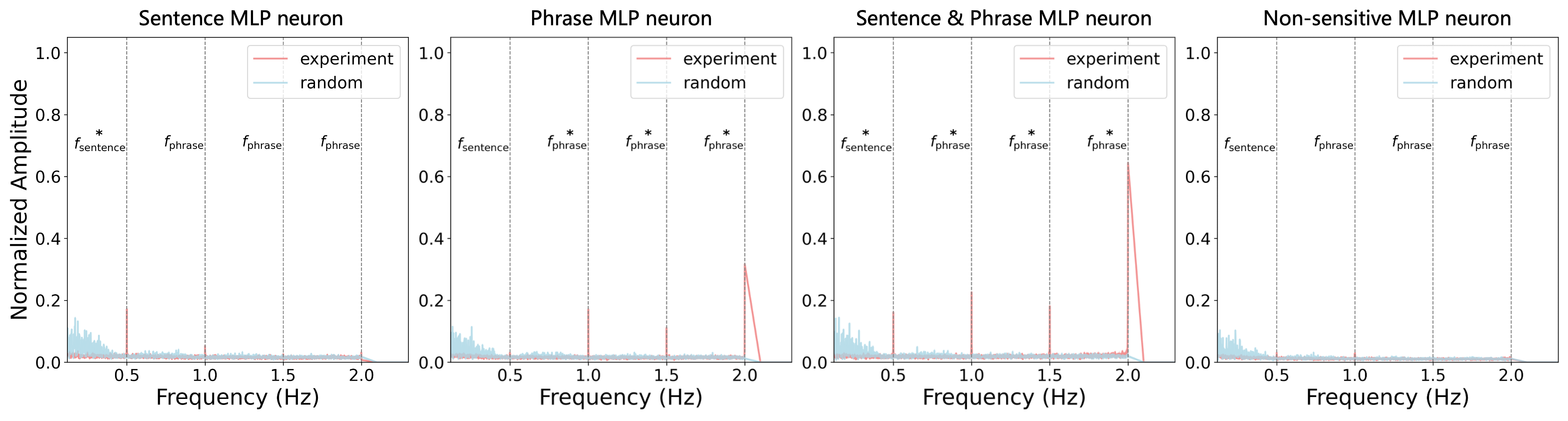}
    \caption{Hierarchical frequency patterns of MLP neurons, using a Chinese Wikipedia 8-syllable corpus, selectively represent sentence features, phrase features, shared features of both, and non-sensitive features (from left to right).}
    \label{naturalwic}
\end{figure}

\begin{figure}[ht]
    \centering
    \includegraphics[width=\linewidth]{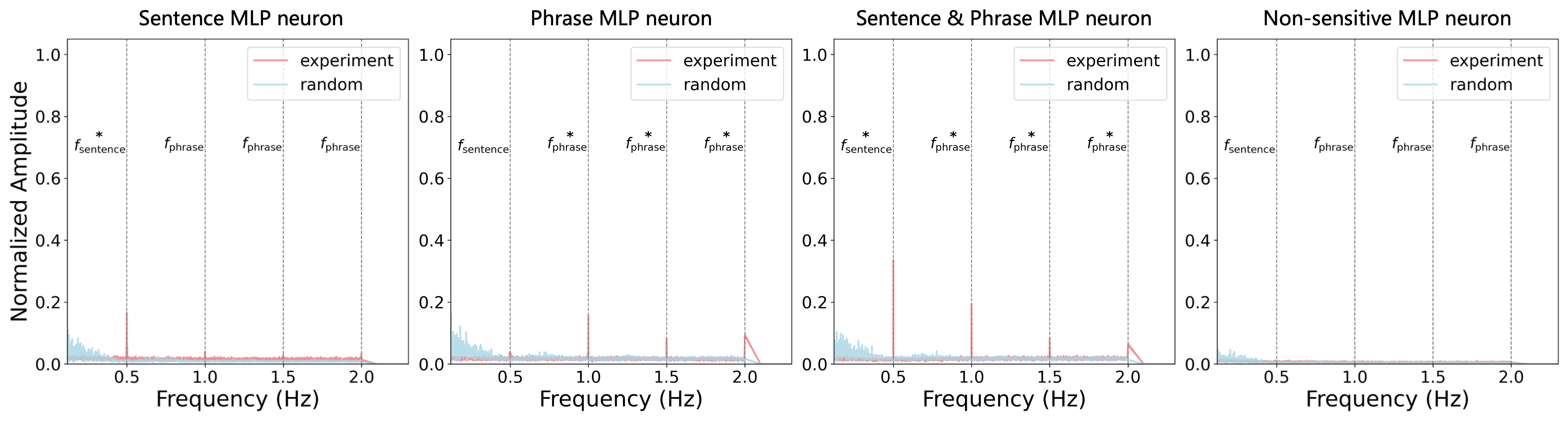}
    \caption{Hierarchical frequency patterns of MLP neurons, using a English Wikipedia 8-word corpus, selectively represent sentence features, phrase features, shared features of both, and non-sensitive features (from left to right).}
    \label{naturalwie}
\end{figure}

We further evaluated HFTP on Chinese and English Wikipedia 8-syllable/word corpora (Figures \ref{naturalwic}–\ref{naturalwie}). In both languages a clear sentence peak appears at \(0.5\;\mathrm{Hz}\), and phrase-rate peaks persist at \(1.0\), \(1.5\), and \(2.0\;\mathrm{Hz}\); however, relative to non-Wikipedia corpora the peaks are attenuated, the separation between experiment and random controls is smaller, and the intermediate \(1.5\;\mathrm{Hz}\) component is more variable.  Although sequence length was strictly controlled to eight syllables/words, Wikipedia text is less regular in content: mixed scripts/orthographies (e.g., simplified vs. traditional in Chinese), frequent abbreviations and alphanumeric tokens, bibliographic fragments and formulaic titles, heterogeneous named entities, inconsistent prosodic phrasing, etc. These factors reduce cross-sentence periodicity and weaken harmonic reinforcement, thereby weakening tracking at sentence- and phrase-rate rhythms.

We speculate that the four dominant frequencies arise from tokenisation statistics and prosodic templating shared across languages: frequent two- and three-token words and binomial/trinomial chunks in English, and abundant 2-/3-character words and 4-character idioms in Chinese; both yield stable 2- and 3-beat groupings, while clause-level phrasing produces a half-sentence rhythm.  Transformer layers further reinforce harmonics of the basic word cycle; in the nine-syllable corpus, a 3-beat unit around \(1.33\;\mathrm{Hz}\) naturally gives rise to a higher harmonic near \(1.78\;\mathrm{Hz}\). The half-sentence peak at \(0.89\;\mathrm{Hz}\) emerges because speakers often place a prosodic break near the midpoint of nine-character or nine-word clauses, creating a stable sub-sentence rhythm that the model entrains to. Because the underlying rhythmic structure is shared, we believe HFTP is highly likely to generalise to other character-centric languages (Japanese, Korean) and to space-delimited alphabetic languages (French, German).

\section{Model details}
\label{model_detail}
In this appendix, we present the details of the LLMs used in this study. Table \ref{model_description} summarizes key parameters, including model size, number of layers, attention heads, and MLP neurons.

\begin{table}[h]
\centering
\caption{Comparison of model parameters.}
\label{model_description}
\begin{tabular}{ccccc}
\toprule
Model& Size& Layer& Attention head& MLP neuron\\
\midrule
GPT-2 \citep{radford2019language}&         774M&         36&         20&         5120\\
Gemma \citep{team2024gemma}&         7B&         28&         16&         24576\\
Gemma 2 \citep{team2024gemma2}&         9B&         42&         16&         14336\\
Llama 2 \citep{touvron2023llama}&         7B&         32&         32&         11008\\
Llama 3.1 \citep{dubey2024llama}&         8B&         32&         32&         14336\\
 GLM-4 \citep{glm2024chatglm}& 9B& 40& 32&13696\\
\bottomrule
\end{tabular}
\end{table}

\section{Syntactic corpora}
\label{corpus}
For our HFTP experiments, we employed six corpora: two syntactic sets—a Chinese four-syllable corpus (Table \ref{corpus_chinese}) and an English four-word corpus (Table \ref{corpus_english})—adapted from \citep{dingCorticalTrackingHierarchical2016}; two human-generated naturalistic Chinese corpora with eight and nine syllables (Tables \ref{corpus_natural1} and \ref{corpus_natural2}); and two parallel naturalistic English corpora with eight and nine words (Tables \ref{corpus_natural8e} and \ref{corpus_natural9e}). These corpora were used to assess model sensitivity to hierarchical structure. In addition, we constructed Wikipedia-derived Chinese and English 8-syllable/word corpora for out-of-domain validation; these corpora are available on \url{https://github.com/LilTiger/HFTP}.

\begin{CJK*}{UTF8}{gbsn} % 开始中文环境，使用UTF-8编码和简体中文字体

\begin{table}
\centering
\caption{Chinese syntactic corpus.}
\label{corpus_chinese}
\begin{tabular}{l l l l l}
\toprule
\multicolumn{5}{c}{Four-syllable sequences}\\
\midrule
老牛耕地 & 朋友请客 & 厨师做饭 & 竹鼠啃笋 & 农民种菜 \\
青草发芽 & 和尚念经 & 老师讲课 & 鲸鱼喷水 & 绵羊吃草 \\
英雄救火 & 游客爬山 & 鸭子划水 & 蜘蛛结网 & 祖父下棋 \\
医生看病 & 护士打针 & 母鸡下蛋 & 行人过街 & 法官判案 \\
狮子吃肉 & 老鹰捕鱼 & 蜜蜂采花 & 小孩读书 & 司机开车 \\
画家作画 & 船夫摇桨 & 诗人吟诗 & 麻雀筑巢 & 猴子摘桃 \\
渔夫撒网 & 骆驼饮水 & 狐狸捕鼠 & 海豹顶球 & 小猫抓鱼 \\
老马拉车 & 鸽子衔枝 & 孩童拾贝 & 雏鸡啄米 & 山雀捉虫 \\
青鸟啄木 & 樵夫砍柴 & 黑熊爬树 & 土狼挖洞 & 军鸽传信 \\
燕雀喂仔 & 野猪拱地 & 渔民划船 & 蚯蚓钻土 & 蚕蛾吐丝 \\
\bottomrule
\end{tabular}
\end{table}

\end{CJK*} % 结束中文环境

\begin{table}
\centering
\caption{English syntactic corpus.}
\label{corpus_english}
\begin{tabular}{l l l}
\toprule
\multicolumn{3}{c}{Four-word sequences}\\
\midrule
fat rat sensed fear & wood shelf holds cans & tan girls drove trucks \\
gold lamps shine light & dry fur rubs skin & sly fox stole eggs \\
top chefs cook steak & our boss wrote notes & two teams plant trees \\
all moms love kids & new plans give hope & large ants built nests \\
teen apes hunt bugs & rude cats claw dogs & rich cooks brewed tea \\
fun games waste time & pink toys hurt girls & huge waves hit ships \\
deaf ears hear you & his aunt tied shoes & kind words warm hearts \\
long fight caused hate & dead sharks leak blood & smart dogs dig holes \\
slim kids like jeans & sick boys fail tests & rear doors hide cups \\
pale hands make bread & bad smells fill town & mad foes smack chefs \\
quiet lamb ate grass & soft fork brings food & green frogs miss flies \\
black skies show stars & tall guys flee camp & gray goat climb hills \\
iced beer costs cents & old kings gave speech & blue eyes shed tears \\
white cars need gas & young child closed doors & thin threads hang plates \\
their store sold cars & cute cubs drink milk & six farms lost cows \\
sharp knife cuts cheese & round soap killed germs & loud sound scared mom \\
weird clowns wear hats & her sons paint walls &  \\
\bottomrule
\end{tabular}
\end{table}

\begin{CJK*}{UTF8}{gbsn} % 开始中文环境

\begin{table}
\centering
\caption{Naturalistic Chinese syntactic corpus.}
\label{corpus_natural1}
\begin{tabular}{l l l}
\toprule
\multicolumn{3}{c}{Eight-syllable sequences}\\
\midrule
森林火势得到控制。     & 列车准点抵达站台。     & 今晚，一起看电影吗？   \\
中央推出税收优惠。     & 请稍等，我来核对价。   & 空调滤网需要更换。     \\
研究论文成功发表。     & 画展现场气氛静谧。     & 港口货轮密集靠泊。     \\
明天，早饭想吃啥呢？ & 请问洗手间在哪呀？     & 他缓缓走入雨巷中。   \\
古桥石栏苔痕深留。   & 图书馆今天人较多。   & 教育部发布新课程。     \\
快看，雨停了出门吧！ & 市场需求逐步回暖。   & 点单吗？我们有套餐。 \\
快递包裹正在派送。   & 图书销量榜单更新。   & 雨夜街灯映成河影。   \\
证券监管再次收紧。   & 服务员，来两杯绿茶！ & 警方破获网络诈骗。   \\
同学，你借我下笔吧？ & 科研团队揭量子谜。   & 春雷滚过江南田畔。   \\
旅行社推出特价游。   & 学生提交毕业论文。 & 他戴上耳机工作中。   \\
老师，这题怎么写呢？ & 他整理旧照片回忆。 & 海浪轻拍沙滩细岸。   \\
亲爱的，晚餐想吃啥？ & 喂，你现在到哪里了？ & 股市午盘震荡收高。   \\
你好，咖啡需要糖吗？ & 桂花香飘整条街上。 & 日出染红东海天际。   \\
月光洒落在城墙上。   & 乡村集市热闹开张。 & 记者现场连线报道。   \\
航班因雾全部延误。   & 数据中心全面升级。 & 雨天，道路易积水患。   \\
智能巴士全线运营。   & 剧院上演经典芭蕾。 & 博物馆新增夜场票。   \\
南部遭遇强降雨灾。   & 想不到你如此嚣张。 &                      \\
\bottomrule
\end{tabular}
\end{table}

\begin{table}
\centering
\caption{Naturalistic Chinese syntactic corpus.}
\label{corpus_natural2}
\begin{tabular}{l l l}
\toprule
\multicolumn{3}{c}{Nine-syllable sequences}\\
\midrule
临床试验数据公布了。   & 姐姐，这裙子有蓝色吗？ & 软件更新，漏洞已修复。 \\
晚上一起吃寿司如何？ & 孩子，慢点吃别噎到啊。 & 付款可用微信，是不是？ \\
医护人员彻夜守病房。 & 智能家居系统升级中。 & 校园社团招募新成员。 \\
我们有可乐、果汁、豆浆。 & 高中将设人工智能课。 & 你好，考试时间改了吗？ \\
摄影展聚焦城市微光。   & 他翻身查看夜空星图。 & 街角花店玫瑰已售罄。 \\
师傅，去火车站多少钱？ & 大雨突临，烟花秀取消。 & 他低声读完那封旧信。 \\
科研数据平台上线啦。   & 社区篮球赛今晚开哨。 & 市场监管局突击检查。 \\
木星再添两颗小卫星。   & 疏影横斜暗上书窗敲。 & 晚餐菜单更新完毕了。 \\
明早七点机场见，好吗？ & 老板，这条鱼再便宜点？ & 卫星成功捕捉极光影。 \\
雨滴敲击玻璃声清脆。   & 电商推广使用绿色包。 & 他轻扣门板等待回应。 \\
喂，你到公司门口了吗？ & 旅行箱在传送带循环。 & 服务员，账单麻烦拿来。 \\
凌晨街头灯火渐稀少。   & 国家队再夺巴黎首金。 & 音乐渐缓舞步更轻盈。 \\
早高峰地铁挤满乘客。   & 新剧首播口碑节节攀。 & 灯光映照湿润青石路。 \\
医生，这药饭前还是后？ & 请坐，这里视线最好哦。 & 你好，请填写到访登记。 \\
咖啡豆飘散焦糖香气。 & 请稍候，我去取电影票。 & 他轻敲键盘修改代码。 \\
图书销量排行榜刷新。   & 同学，笔记本借我一下？ & 志愿者分发食物物资。 \\
快递无人签收被退回。   & 博主分享无人机航拍。 &                      \\
\bottomrule
\end{tabular}
\end{table}

\end{CJK*} % 结束中文环境

\begin{table}[t]
\centering
\caption{Naturalistic English 8-word corpus.}
\label{corpus_natural8e}
\small
\setlength{\tabcolsep}{3pt}
\begin{tabular}{@{}>{\raggedright\arraybackslash}p{0.48\linewidth} >{\raggedright\arraybackslash}p{0.48\linewidth}@{}}
\toprule
\multicolumn{2}{c}{Eight-word sequences}\\
\midrule
With malice toward none, with charity toward all. & Tonight, shall we watch the meteor shower together? \\
A man can be destroyed but not defeated. & Excuse me, where's the nearest restroom around here? \\
The library felt crowded during rainy examination week. & Market demand seems gradually rebounding during this quarter. \\
Book sales rankings were refreshed early this morning. & Waiter, two cups of hot green tea please. \\
Quantum research team unveiled perplexing entanglement results yesterday. & Students submitted their final graduate theses by noon. \\
He sorted aging photographs, reminiscing about bygone days. & Hello, where are you right now, my friend? \\
Moonlight spilled gently across the ancient city walls. & Flights were delayed nationwide due to dense fog. \\
Autonomous buses now operate on every urban route. & Southern region suffered severe flooding after relentless rain. \\
Plum blossoms scented the courtyard with delicate sweetness. & Please hold on, I'll check your booking details. \\
Central bank announced fresh stimulus to boost economy. & Port cranes unloaded containers under brilliant afternoon sky. \\
He slowly wandered into the rainy narrow alleyway. & Education ministry announced updated national curriculum guidelines today. \\
Ready to order? We have today's special combo. & Rainy night streetlights shimmered like a glowing river. \\
Police dismantled an extensive online fraud network operation. & Spring thunder rolled over rice paddies in Jiangnan. \\
He put on headphones and focused on coding. & Gentle waves lapped softly against the sandy shoreline. \\
Could you use some sugar in your coffee? & The rural marketplace opened lively at dawn today. \\
The data center completed a comprehensive system upgrade. & The theater staged a timeless classical ballet tonight. \\
I never imagined you could be this arrogant. & City skyline glittered beneath a crisp winter moon. \\
Train whistle echoed across fields drenched with mist. & Research findings published in a prestigious journal today. \\
Tomorrow, what would you like for breakfast, friend? & Ancient stone bridge retained mossy grooves through centuries. \\
Look, the storm passed; let's explore the streets. & Parcel delivery is currently out for neighborhood distribution. \\
Financial regulators tightened oversight on speculative securities trading. & Hey classmate, may I borrow your pen briefly? \\
Travel agency launched discounted spring break tour packages. & Teacher, could you explain this problem once more? \\
Darling, what would you like for dinner tonight? & Osmanthus fragrance drifted along the entire narrow street. \\
Reporter delivered live coverage from bustling downtown square. & On rainy days, roads easily accumulate dangerous puddles. \\
Museum introduced extended evening hours for visitors' convenience. & Night market vendors grilled skewers over glowing charcoal. \\
\bottomrule
\end{tabular}
\end{table}

\begin{table}[t]
\centering
\caption{Naturalistic English 9-word corpus.}
\label{corpus_natural9e}
\small
\setlength{\tabcolsep}{3pt}
\begin{tabular}{@{}>{\raggedright\arraybackslash}p{0.48\linewidth} >{\raggedright\arraybackslash}p{0.48\linewidth}@{}}
\toprule
\multicolumn{2}{c}{Nine-word sequences}\\
\midrule
Clinical trial data were publicly released this morning nationwide. & Shall we share sushi together tonight by the river? \\
War is peace, freedom is slavery, ignorance is strength. & We offer cola, soda, juice, and soy milk today. \\
Photography exhibition focuses on city's hidden pockets of light. & Driver, how much is the fare to the station? \\
Scientific data platform finally went live to the public. & Jupiter just gained two additional tiny moons last week. \\
See you at the airport tomorrow seven sharp, alright? & Raindrops drummed against windowpanes in a crisp cadence tonight. \\
Hey, have you reached the company entrance yet today? & Early morning city lights gradually faded as traffic intensified. \\
Morning rush subway overflowed with restless hurried commuters again. & Doctor, should this new medicine be taken before dinner? \\
Coffee beans release subtle caramel aroma throughout the cafe. & Book sales leaderboard refreshed, surprising many independent authors today. \\
Undelivered parcels were returned after nobody signed for them. & Sister, does this blue dress come in medium size? \\
Child, slow down and chew, avoid choking on food. & Smart home automation system is currently installing new firmware. \\
High school will introduce artificial intelligence elective next year. & He rolled over, consulting a star chart at midnight. \\
Sudden downpour forced cancellation of tonight's fireworks display completely. & Community basketball tournament tips off under bright evening floodlights. \\
Slanting shadows tapped an old window softly at dusk. & Boss, could this fish be a little cheaper please? \\
Ecommerce campaign now promotes sustainable green packaging materials nationwide. & Luggage carousel kept circling the unattended silver suitcase indefinitely. \\
National team claimed its first gold medal in Paris. & New drama premiere received soaring reviews across social media. \\
Please sit here; the view remains the best available. & Please wait, I will retrieve our cinema tickets now. \\
Classmate, may I borrow your notebook for a moment? & Influencer shared breathtaking drone footage of mountain sunrise yesterday. \\
Software update completed, critical vulnerabilities have been fixed already. & Payment through WeChat is available, is that alright sir? \\
Campus club seeks enthusiastic freshmen to join this semester. & Hello, has the final exam schedule been changed recently? \\
Corner flower shop sold out all roses before noon. & He whispered while finishing that fading wartime love letter. \\
Market supervision bureau conducted an unexpected compliance inspection today. & Dinner menu has been fully updated for the evening. \\
Satellite captured brilliant aurora images above polar orbit yesterday. & He gently knocked, waiting patiently for someone to respond. \\
Please bring the bill, waiter, we are finished here. & Music slowed, dancers embraced lighter steps beneath dimmed lights. \\
Lantern glow reflected onto slick cobblestones after evening drizzle. & Hello, kindly complete the visitor registration form at reception. \\
He tapped the keyboard softly, refining his source code. & Volunteers distributed food supplies to displaced families after flood. \\
\bottomrule
\end{tabular}
\end{table}

For the HFTP experiment in the human brain, we utilized two Chinese corpora: the \textbf{sentence} and \textbf{phrase} corpora. To ensure consistent analysis of syntactic processing across both LLMs and the human brain, the same corpora were applied to the alignment experiment. These corpora originated from \citep{sheng2019cortical}. Participants received a brief oral instruction to listen carefully to each stimulus, completed 40 trials per condition (sentence, phrase, and random), and were compensated approximately \textyen{}200 for full participation, with pay prorated if they completed only a subset of trials.

\section{Brain ROIs}
\label{region_label}

As discussed in Section \ref{process_human}, we reorganized the original sEEG data by grouping the Automated Anatomical Labeling (AAL) annotations into newly defined brain ROIs for our experiments  \mbox{\citep{rolls2015implementation}}. In this appendix, we provide the full names of AAL regions, the corresponding AAL labels used in the sEEG data, and their mapped brain ROIs in Table \ref{regions}.

\begin{table}[ht]
\caption{Automated Anatomical Labeling (AAL) annotations from the original sEEG data, along with their mapped brain ROIs. Note that the regions are distinguished by the left and right hemispheres.}
\label{regions}
\begin{center}
\begin{tabular}{ccc}
\toprule
\multicolumn{1}{c}{\bf AAL Full Name} &\multicolumn{1}{c}{\bf AAL Label}  &\multicolumn{1}{c}{\bf ROI}
\\ \hline
Heschl Gyrus &Heschl           &A1 \\
Superior Temporal Gyrus &Temporal\_Sup    &STG \\
Middle Temporal Gyrus &Temporal\_Mid    &MTG \\
Inferior Temporal Gyrus &Temporal\_Inf    &ITG \\
Parahippocampal Gyrus &ParaHippocampal  &ITG \\
Fusiform Gyrus &Fusiform         &ITG \\
Insular Cortex &Insula           &Insula \\
Angular Gyrus &Angular          &TPJ \\
Supramarginal Gyrus &SupraMarginal    &TPJ \\
Inferior Parietal Lobule &Parietal\_Inf    &TPJ \\
Superior Temporal Pole &Temporal\_Pole\_Sup &Temporal\_Pole \\
Middle Temporal Pole &Temporal\_Pole\_Mid &Temporal\_Pole \\
Paracentral Lobule &Paracentral\_Lobule &Sensorimotor \\
Supplementary Motor Area &Supp\_Motor\_Area &Sensorimotor \\
Rolandic Operculum &Rolandic\_Oper   &Sensorimotor \\
Precentral Gyrus &Precentral       &Sensorimotor \\
Postcentral Gyrus &Postcentral      &Sensorimotor \\
Inferior Frontal Gyrus, Opercular part &Frontal\_Inf\_Oper &IFG \\
Inferior Frontal Gyrus, Triangular part &Frontal\_Inf\_Tri &IFG \\
Inferior Frontal Gyrus, Orbital part &Frontal\_Inf\_Orb &IFG \\
Middle Frontal Gyrus &Frontal\_Mid     &MFG \\
Middle Frontal Gyrus, Orbital part &Frontal\_Mid\_Orb &MFG \\
Hippocampus &Hippocampus      &Hippocampus \\
Amygdala &Amygdala         &Amygdala \\
\bottomrule
\end{tabular}
\end{center}
\end{table}

\end{document}